\documentclass[sigconf]{acmart}

\usepackage{multirow}
\usepackage{enumitem}
\usepackage{makecell}
\usepackage{arydshln}
\usepackage{mdframed}
\usepackage{balance}

\newcommand{\miniskip}{\vspace*{-.5\baselineskip}}
\newcommand{\shrink}{\vspace*{-.9\baselineskip}}

\AtBeginDocument{%
  }

\copyrightyear{2024}
\acmYear{2024}
\setcopyright{rightsretained}
\acmConference[SIGIR-AP '24]{Proceedings of the 2024 Annual International ACM SIGIR Conference on Research and Development in Information Retrieval in the Asia Pacific Region}{December 9--12, 2024}{Tokyo, Japan}
\acmBooktitle{Proceedings of the 2024 Annual International ACM SIGIR Conference on Research and Development in Information Retrieval in the Asia Pacific Region (SIGIR-AP '24), December 9--12, 2024, Tokyo, Japan}\acmDOI{10.1145/3673791.3698415}
\acmISBN{979-8-4007-0724-7/24/12}

\begin{document}

\title{Fine Tuning vs. Retrieval Augmented Generation for Less Popular Knowledge}

\author{Heydar Soudani}
\affiliation{%
  \institution{Radboud University}
  \city{Nijmegen}
  \country{The Netherlands}}
\email{heydar.soudani@ru.nl}

\author{Evangelos Kanoulas}
\affiliation{%
  \institution{University of Amsterdam}
  \city{Amsterdam}
  \country{The Netherlands}}
\email{e.kanoulas@uva.nl}

\author{Faegheh Hasibi}
\affiliation{%
 \institution{Radboud University}
 \city{Nijmegen}
 \country{The Netherlands}}
\email{faegheh.hasibi@ru.nl}

\renewcommand{\shortauthors}{Heydar Soudani, Evangelos Kanoulas, \& Faegheh Hasibi}

\begin{abstract}
Language Models (LMs) memorize a vast amount of factual knowledge, exhibiting strong performance across diverse tasks and domains. However, it has been observed that the performance of these models diminishes when dealing with less-popular or low-frequency concepts, for example, in domain-specific applications. The two prominent approaches to enhance the performance of LMs on less frequent topics are Retrieval Augmented Generation (RAG) and fine-tuning (FT) over synthetic data. This paper explores and evaluates the impact of RAG and FT on customizing LMs in handling low-frequency entities in question answering tasks. We conduct extensive experiments on twelve LMs of varying size and type and different FT methods, data augmentation, and retrieval models. Our findings indicate that while FT boosts performance across entities of varying popularity, RAG surpasses FT by a large margin, particularly for least popular factual knowledge. Additionally, the success of both RAG and FT approaches is amplified by improving retrieval and data augmentation techniques. Fine-tuning, while beneficial for small LMs, requires extensive resources. To address this issue, we propose the new \emph{Stimulus RAG} approach that surpasses the effectiveness of fine-tuning based approaches, thereby eliminating the need for the costly data augmentation and fine-tuning steps for enriching LMs with less popular factual knowledge. 
The code is available at \url{https://github.com/informagi/RAGvsFT}.
\end{abstract}





\maketitle

\section{Introduction}~\label{sec:intro}

\begin{figure}[t]
  \centering
  \includegraphics[width=0.49\textwidth]{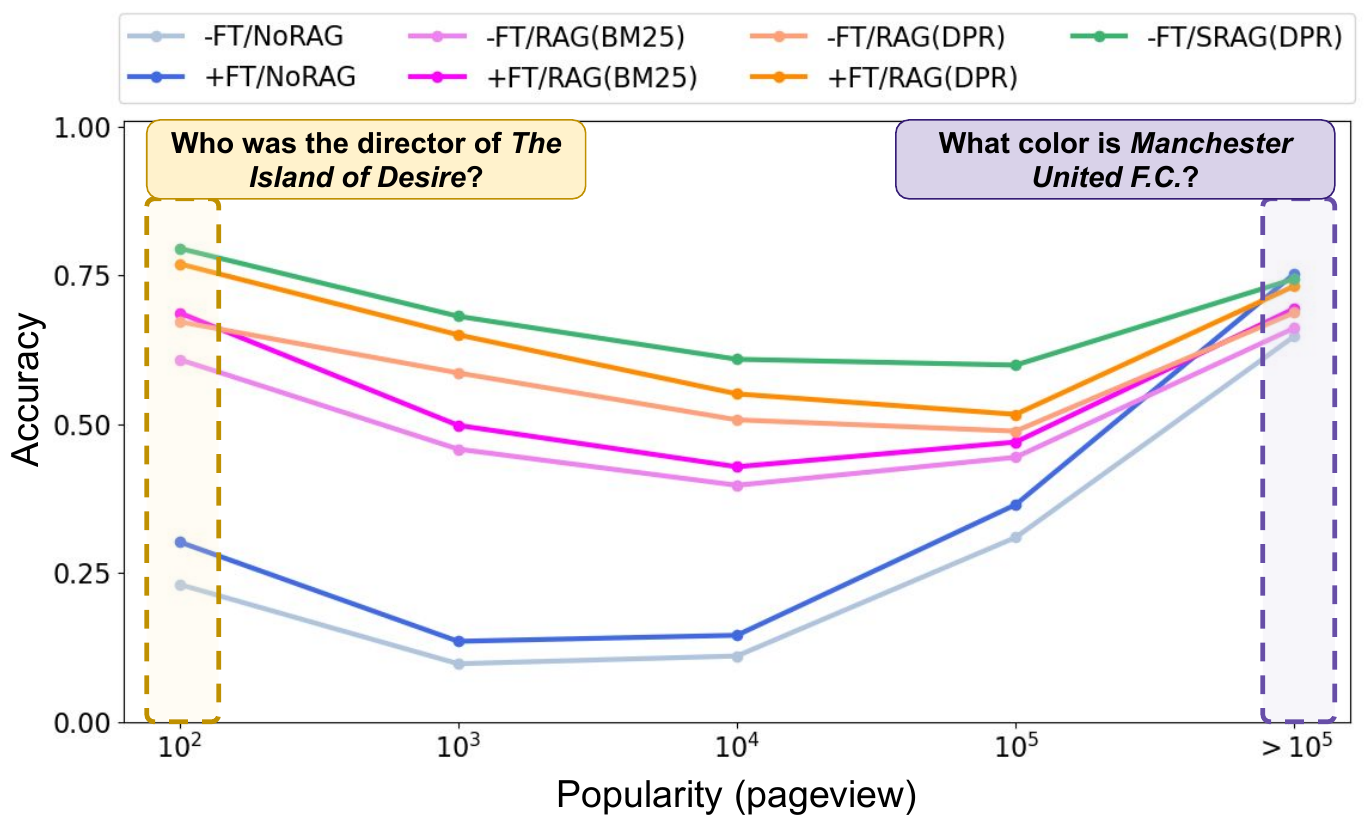}
  \shrink

  \caption{Comparison of RAG and fine-tuning 
  on StableLM2 performance in question answering over factual knowledge. RAG-based approaches significantly enhance the performance of the vanilla StableLM2, outperforming fine-tuning by a large margin. Our proposed SRAG approach outperforms all models, including the fine-tuning based approaches.}
  % ccompared to Popularity of subject entities is measured by page views, demonstrated   Correlation between subject entity popularity in a question and the effects of RAG and FT on StableLM2 performance in open-domain question answering. FT markedly improves accuracy in the initial and final buckets relative to others (indicated by the blue line).}
  \label{fig:main_fig}
  \shrink
\end{figure}

% == Adapting a pre-trained model for the specific application
% = The problem is less-popular entities
Language Models (LMs) exhibit outstanding capabilities in executing tasks that demand extensive memorization of factual data~\citep{PaLM23Chowdhery}. However, their memorization capabilities are constrained when dealing with less frequent entities~\cite{When23Mallen, Gerritse:2022:EMBERT, Large23Kandpal, Head23Kai}, and even the largest models may encounter the well-known "hallucination" problem~\cite{Retrieval21Shuster} and temporal degradation~\cite{RealTime22Kasai}.
Consequently, when LMs are intended for deployment in less resourced domains, customization becomes imperative to ensure optimal performance. A common example is within the industrial setup, where chatbots or Question Answering (QA) systems need to accurately answer users' questions about a proprietary knowledge graph or intra-company terminology with limited textual description~\cite{Zhang24raft, Data23soudani}.

% == RAG & FT are two solutions
Retrieval-Augmented Generation (RAG) and Fine-Tuning (FT) stand out as two prominent approaches for adapting LMs to specific domains~\cite{Data23soudani, Few23Mosbach, RAG24Balaguer, Fine23Ovadia}. RAG retrieves relevant information from a document corpus and enhances LM's response generation through the implementation of in-context learning (ICL)~\cite{Zhang24raft, Noise2024Florin}. Conversely, FT approach updates model weights to become adept at recalling specific information and enhance its memorization capabilities during inference~\cite{Reliable24Asai}. In the context of less popular knowledge, where limited data is available, data augmentation methods are utilized to generate synthetic training data, serving as an initial step towards FT~\cite{survey24soudani, Augmentation24soudani}. 
Despite existing research on enhancing LM's memorization with RAG~\cite{When23Mallen}, no work to our knowledge has compared RAG with  with knowledge obtained through FT, particularly for less popular knowledge.% \todo{MAybe we can here something along the lines that no existing study compare these two approaches?}
% \new{Although the effect of RAG has been compared with memorized knowledge in LMs~\cite{When23Mallen}, to the best of our knowledge, no work has compared it with knowledge obtained through FT, particularly for less popular knowledge.}
% 

% == Our contributions and RQs

% ==== V1:
% \begin{enumerate}[label=(RQ\arabic*)]
%     \item What is the effectiveness of RAG and fine-tuning with synthetic data on QA for low-frequency factual knowledge?
%     \item Which parameters, including the quality of synthetic samples, the method to be fine-tuned, the model size, and the performance of retrieval models affect the downstream performance?
%     \item Can we bypass FT with a more advanced RAG system?
% \end{enumerate}
% ==== V2:
% \begin{enumerate}[label=\textbf{(RQ\arabic*)}]
%     \item~\label{rq:1} To what extent can factual knowledge be memorized by LMs, how can the memorization process be optimized considering available resources, and what factors affect this memorization?
%     \item~\label{rq:2} How does the performance of retrieval-augmented LMs compare to non-customized and customized LMs?
%     \item~\label{rq:3} Can a more advanced RAG system be built to obtain the advantages of fine-tuned LMs while bypassing the efforts required for FT?
% \end{enumerate}
% ==== V3:
In this paper, we aim to understand which approach is more appropriate for customizing LMs for less resourced domains. Specifically, we seek to answer this research question: \emph{\textbf{(RQ1)} How does RAG compare to fine-tuning for question answering over less popular factual knowledge, and which factors affect their performance?}
To address this question, we conduct a comprehensive comparison of RAG and fine-tuning methods for less popular knowledge, assuming that textual descriptions, albeit limited, are available for a specific domain and application. We, therefore, collect Wikipedia documents related to QA datasets over long-tail entities and apply two methods of knowledge injection: parametric knowledge injection using FT and non-parametric knowledge injection using RAG. For the FT approach, the LM is fine-tuned with synthetically generated QAs from these documents using data augmentation approaches~\cite{Synthetic19Alberti, Empirical2023ushio}. For the RAG approach, we use retrievers to rank the most relevant documents for a query. 
%The generated QAs are injected into the LMs as parametric knowledge, while the top-ranked documents are directly inserted into the input prompt as non-parametric knowledge. 
We investigate how the effectiveness of these methods is affected by the following aspects: (i) fine-tuning method; i.e., full FT vs. parameter efficient fine-tuning (PEFT), (ii) data augmentation method, (iii) LM type and size; i.e., decoder only vs. encoder-decoder models and varying size, ranging from 80M to 11B parameters, and (iv) retrieval model performance. 
Through exhaustive experimentation on twelve LMs and different setup of fine-tuning, data augmentation, and retrieval models, we arrive at the following conclusions:
\shrink
\begin{itemize}
    \item \textbf{Fine-tuning method:}
    % \todo{Summarize our findings in a couple of sentences}
    Comparing full FT with PEFT (i.e., QLoRA~\citep{QLoRA23Dettmers}) for LMs with less than 2 billion parameters, full FT is more effective than PEFT in the downstream task. PEFT, however, preserves the reasoning ability of LMs (needed for RAG) and outperforms full FT when the fine-tuned models are used in combination with RAG (Table~\ref{tb:ft_qa_results}).
    % the but less efficient in terms of training. PEFT, while being competitive to full FT, maintains the reasoning ability of LMs used in  
    
    \item \textbf{Data augmentation method:} 
    % \todo{Summarize our findings in a couple of sentences}
    Comparing prompt-based and a state-of-the-art fine-tuned~\cite{Empirical2023ushio} QA generation models, the prompt-based method demonstrates better performance for the downstream task. This suggests that the high-quality synthetic data generated by large LMs can better assist LMs with memorizing new knowledge, compared to the greater volume of data generated by the fine-tuned model (Table~\ref{tb:ft_qa_results}).
    % have a superior ability to generate high-quality QA pairs, although the fine-tuned model produces a greater volume of data 
    
    \item \textbf{LM type and size:}
    % \todo{Summarize our findings in a couple of sentences}
    % Additionally, fine-tuning encoder-decoder LMs is more time-consuming. 
    % Furthermore, increasing the size of LMs improves their performance before fine-tuning. 
    Comparing  decoder-only with encoder-decoder LMs (Flan-T5 models of various sizes), decoder-only models outperform encoder-decoder models of similar size. Interestingly, larger LMs generally do not benefit from fine-tuning, while smaller ones do. Therefore, a small fine-tuned LM with RAG can perform on par or better than a large LM; e.g., StableLM2 (1.6B) vs. Llama3 (8B) (Table~\ref{tb:all_results}).
    
    \item \textbf{Retrieval model:}
    Comparing retrievers with varying performance in the RAG system, we observe that as the popularity of factual knowledge increases, the performance of the retriever decreases (Figure~\ref{fig:ret_recall}).  Moreover, the performance of the RAG system increases by using higher performance retriever (Figures~\ref{fig:main_fig} and \ref{fig:ret_res_bk}).
    %By investigating the effectiveness of the retrievers with varying performance in the RAG system, our observations show that as the popularity of factual knowledge increases, the performance of the retriever decreases (Figure~\ref{fig:ret_recall}). This indicates that for popular concepts, the retrieval-augmented prompt contains noise. Interestingly, for very popular entities, LMs are able to instinctively ignore the input prompt and use their internally embedded knowledge. Moreover RAG's performance increases by using more higher performance retriever (Figures~\ref{fig:main_fig} and \ref{fig:ret_res_bk}).
    
    \item \textbf{Fine-tuning vs. RAG:} 
    Comparing these two knowledge injection methods, RAG substantially outperforms fine-tuning. Fine-tuned LMs combined with RAG  either outperform or perform on par with vanilla LMs with RAG in all but one case (Figure~\ref{fig:main_fig}).
\end{itemize}

While fine-tuning improves accuracy in answering factual questions, both with and without RAG, it demands a considerable amount of effort and resources. This leads us to our second research question: \emph{\textbf{(RQ2):}  Can we avoid the cost of fine-tuning by developing an advanced RAG approach that surpass the performance of a fine-tuned LM with RAG?}
To answer this question, we develop \emph{Stimulus RAG (SRAG)}, a new RAG approach that stimulates an LM to generate the correct response based on the provided \emph{hint} in the prompt. The hint is extracted from the top retrieved documents by the retrieval model. Our results demonstrate that \emph{Stimulus RAG outperforms all other combinations of fine-tuning, both with and without retrieve-then-generate RAG.}

To summarize, this paper makes the following contributions:

\begin{itemize}
    \item We study the effectiveness of fine-tuning and RAG approaches for question answering over less popular factual knowledge and compare the performance of these models across distinct setups: vanilla and fine-tuned models, both with and without RAG, using different data augmentation methods.
    \item We perform extensive experiments to understand how fine-tuned and RAG models are affected by four different factors: data augmentation method, fine-tuning method, LM type and size, and retrieval model.
    \item We propose a new RAG approach \emph{Stimulus RAG} that outperforms all RAG and fine-tuning setups, thereby bypassing the need for expensive fine-tuning. 
\end{itemize}
% Although FT offers advantages in terms of accuracy on downstream tasks, both with and without RAG, it requires a considerable amount of effort and resources. As a result, we propose a new stimulus RAG approach to achieve the accuracy of a fine-tuned LM without actually fine-tuning it \emph{\textbf{(RQ2)}}. We argue that increasing the number of documents is not helpful and can sometimes mislead the LMs. However, wisely highlighting specific parts of the documents can guide LMs to find the correct answer.

\section{Related Work}~\label{sec:related_work}

\noindent
\textbf{Parametric and Non-parametric Knowledge.}
% 
% 1) Parametric: DA and finetuning
It is demonstrated that large pre-trained LMs memorize a significant amount of world knowledge in their parameters (\textit{parametric knowledge})~\cite{When23Mallen}. FT can update the parametric knowledge embedded in LMs and customize it for a specific domain~\cite{Few23Mosbach, Gerritse:2022:EMBERT}.
One of the important principles for FT is data availability, which is limited specifically in specialized domains~\cite{RAG24Balaguer, Fine23Ovadia}. 
Data augmentation (DA) addresses the data scarcity problem by generating task- and domain-relevant samples from existing unlabeled texts. A common DA approach for the QA task is generating question-answer pairs through a four-step \emph{Pipeline}, consisting of passage selection, answer extraction, question generation, and consistency filtering~\citep{Synthetic19Alberti, Unsupervised19Lewis, PAQ21Lewis, Empirical2023ushio}. 
\citet{Empirical2023ushio} conducted an empirical study comparing three QA generation approaches: Pipeline, Multitask, and End-to-End (E2E) and showed the  E2E approach outperforms others in downstream tasks.

% 2) non-parametric: RAG
Furthermore, a large body of work shows that augmenting LMs with \textit{nonparametric knowledge} (i.e., retrieved text chunks) enables much smaller models to match the performance of larger models~\citep{Retrieval20Patrick}.
In this method, known as Retrieval Augmented Generation (RAG), an information retrieval system is utilized to find relevant documents and adds them to the input prompt to enhance response generation of LMs~\citep{Retrieval23Asai, SelfRAG23Asai}.

% 3) Comparing RAG vs FT
As interest grows in refining pre-trained LMs for particular tasks, the comparison of FT and RAG strategies under equitable conditions is becoming increasingly important.
\citet{Few23Mosbach} explored the effectiveness of few-shot FT versus ICL for classification tasks in general domains. \citet{RAG24Balaguer} compared FT and RAG in answering agriculture and geography-specific questions. \citet{Fine23Ovadia} assessed the performance on multiple-choice questions in specialized areas like anatomy, astronomy, biology, and prehistory. In contrast to these studies, we directly address the integration of less popular factual knowledge into LMs, comparing various retrievers, data augmentation, and fine tuning methods.

\noindent
\textbf{Less Popular Knowledge.}
An entity's popularity in LMs is gauged by its frequency in the model's pre-training data~\citep{Benchmarking23Godbole, Nonparametric23Min}, often assessed through the entity's occurrences in a large corpus~\citep{Large23Kandpal} via entity linking~\citep{Hulst:2020:REL, decao2021autoregressive}. Due to the practical challenges of direct counting, e.g., annotation of large-scale collections with entities~\cite{Kamphuis:2023:MMEAD}, proxies are defined to approximate the popularity of factual knowledge. \citet{Head23Kai} use traffic metrics and content density, while \citet{maekawa24witqa} introduce the co-occurrence of the subject entity and relation predicate as a popularity proxy. Wikipedia pageviews are among the most prevalent methods for measuring the popularity of entities~\citep{When23Mallen, Simple21Sciavolino, Evaluating20Chen}.

\begin{figure}[t]
  \centering
  \includegraphics[width=0.48\textwidth]{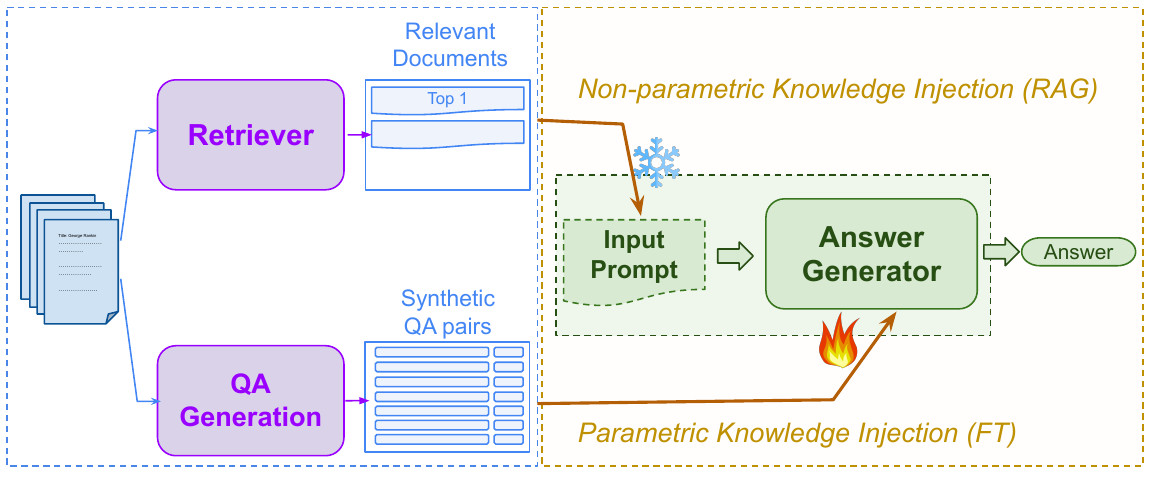}
  \shrink
  \shrink
  \caption{Overview of parametric and non-parametric  knowledge injection for less popular factual knowledge. First, we prepare the corpus. Next, we generate knowledge in two formats: textual documents and synthetic QA pairs. Finally, we inject the knowledge into the prompt or LM parameters.}
  \label{fig:eval_setup}
\shrink
\end{figure}

\noindent
% \textbf{Highlighting Passage.}
\textbf{RAG Development.}
RAG introduces a new approach in AI, combining the strengths of both retrieval-based and generative models~\cite{Noise2024Florin}. The concept of RAG was coined and popularized by \citet{Retrieval20Patrick}, who introduced a model that combines a dense passage retriever with a sequence-to-sequence model, called Retrieve-then-Generate. This approach demonstrated substantial improvements in knowledge-intensive tasks. Several parameters affect a RAG system's accuracy, including the performance of the retriever model~\cite{Lift23Cheng}, the relevance of the passages included in the prompt context, their position, and their number~\cite{Noise2024Florin, Reliable24Asai}.

However, several works have argued that the Retrieve-then-Generate approach is not optimal for more complex tasks, and more advanced RAG systems are needed. Adaptive-RAG~\cite{Jeong24Adaptive} defines a classification-based RAG system to decide which RAG model should be used based on the question type. RAFT~\cite{Zhang24raft} trains a model to ignore documents that don’t help in answering the question, thereby adapting LMs to domain-specific RAG. Generate-then-Retrieve (GR)~\cite{Generate24Abbasiantaeb} argues that the Retrieve-then-Generate paradigm is insufficient when the answer must be obtained from multiple documents. They introduce the GR pipeline, which first generates multiple queries and then retrieves information for the generated queries. \citet{When23Mallen} found that for popular knowledge, using RAG can hurt performance, so they defined an adaptive retrieval system to use retrieval only where it is beneficial.

% === Our contribution -> Highlighting
We discuss that increasing the number of documents is not helpful and may introduce more noise into the LM's input. To address this problem, we propose a new stimulus RAG system that highlights parts of the input text most likely to contain the correct answer. This approach aids the LM in accurately identifying and extracting relevant information.
Highlighting has been used in the literature of information retrieval for various purposes. \citet{Askari23Expand} aim to generate synthetic documents for queries, highlighting keywords to create high-quality documents. \citet{Cho20Better} propose generating sub-sentence summary highlights to overlay on source documents, enabling users to quickly navigate through content. \citet{Li23Guiding} introduce a new prompting framework to provide black-box LMs with fine-grained, instance-specific guidance toward desired outputs. Our work differs from these in that we highlight sentences using a simple yet effective reranker model, which directly improves RAG's performance.

\section{Methodology}~\label{sec:methodology}

In this section, we introduce our evaluation framework (Figure~\ref{fig:eval_setup}), which is designed to assess the effectiveness of two knowledge injection methods: the parametric method using FT and the non-parametric method using RAG. 

\subsection{Task Definition}
This study specifically focuses on factual knowledge~\cite{adams2015bloom} of entities, defined as information about particular attributes and characteristics of target entities, among various types of world knowledge~\cite{When23Mallen}. We chose this because the amount of knowledge memorized by LMs can be approximated by its accuracy in answering simple factual questions, such as "In what city was Lisa Miller born?"~\cite{Head23Kai}.

Factual knowledge is defined as a triplet of (subject, relationship, object)~\cite{When23Mallen}. In this context, the question involves the subject and the relationship, while the answer corresponds to the object. By using these template questions, we ensure that LMs understand the question and derive the answer from their embedded knowledge. 
We select Wikipedia-based question-answering datasets focused on factual knowledge. This enables us to measure the popularity of entities based on Wikipedia pageviews and also  obtain the corresponding evidence document for each entity from Wikipedia.
For each dataset, we select Wikipedia pages whose corresponding entities appear in the test dataset. This setup mirrors real-world industry practices, where entities and their textual descriptions relate to companies' specific internal concepts.
%  Additionally, it is straightforward to obtain the corresponding evidence document for each entity from Wikipedia. Moreover, Wikipedia allows for measuring popularity based on pageviews. Consequently, we selected Wikipedia-based question-answering datasets focused on factual knowledge. 

\subsection{Knowledge Injection with Fine-tuning}~\label{sec:know_inj_ft}
LMs are primarily pre-trained on general domains. To customize LMs for specific knowledge or a particular domain, fine-tuning (FT), also known as parametric knowledge injection, is commonly used. However, FT requires a substantial amount of training samples, which are often unavailable for specific applications, such as those within a company. Data augmentation (DA) offers a solution to this training data shortage. 
To achieve parametric knowledge injection, we fine-tune LMs using synthetically generated question-answer (QA) pairs. Formally, given a set of documents \( D = \{d_1, \ldots, d_n\} \), where $n$ is the number of documents, a QA pair generator $\mathcal{G}_{qa}$
is tasked with generating as many QA pairs as possible:
\begin{equation*}
    \mathcal{G}_{qa}(d_i) = \{(q_i^1, a_i^2), \ldots, (q_i^m, a_i^m)\},
\end{equation*}
% \[ QA_i = P_{qag}(d_i) \]
where $q_i$ and $a_i$ denote a question and answer generated for the document $d_i$,  $m$ is the total number of generated QAs for the document, and the generated set of question-answer pairs is denoted as  $Q_i$. The QA generation method $\mathcal{G}_{qa}$ is applied to all documents $D$, and the generated set $Q = \bigcup_{i=1}^{n} Q_i$ is then used for fine-tuning an LLM, which consists of a set of learnable parameters to predict the probabilities of future or masked tokens.
% \begin{equation*}
%     P_{ft}(\theta_{ft}) \leftarrow P_{pt}\left(Q, \theta_{pt}\right), 
% \end{equation*} 
% where \( \theta_{pt} \) indicates the generator's parameters before fine tuning, and \( \theta_{ft} \) represents the LM's parameters after fine tuning.

%for that do \( QA_i = \{(q_{i1}, a_{i1}), \ldots, (q_{im}, a_{im})\} \), and the value of \( m \) varies for each \( i \). 

\begin{figure}[t]
    \centering
    \includegraphics[width=0.46\textwidth]{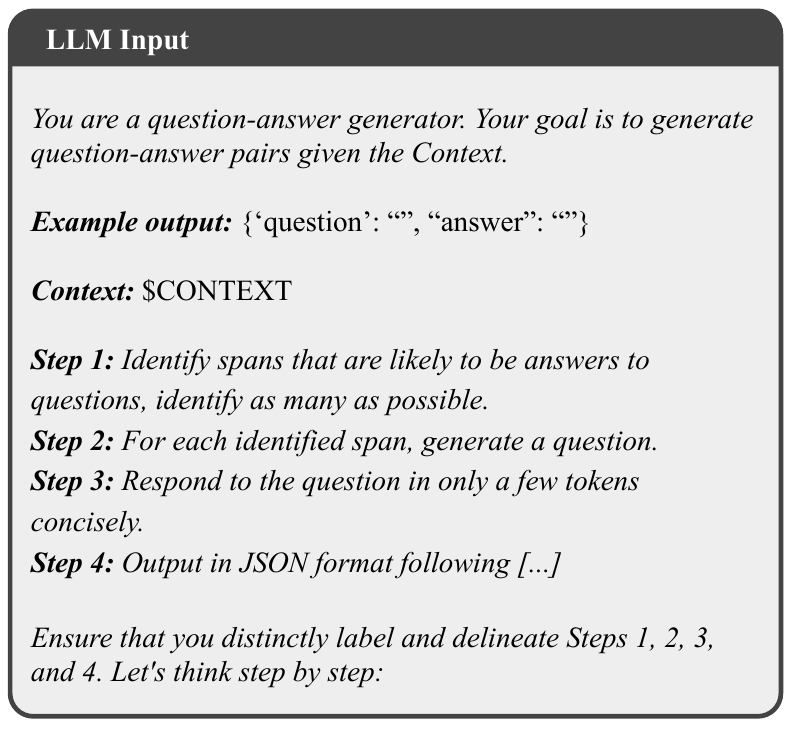}
    \shrink
    \caption{Input prompt for prompt-based QA pair generation. We define a CoT prompt to outline the generation steps. }
    \label{fig:qa_gen_input_generation}
    \shrink
\end{figure}
% === Prompting & E2E method
We employ two QA generation methods. The first one is end-to-end (\textbf{E2E}) QA generation, where a fine-tuned sequence-to-sequence model generates QA pairs from \( d_i \). We utilize the E2E approach by ~\citet{Empirical2023ushio}, which employs a trained T5-large for paragraph-level QA generation and has been shown to be more robust and effective than the established pipeline approach. The term "E2E" is used because the QA generation process is not divided into two sequential components, i.e., answer extraction and question generation. Instead, a QA pair is generated in one go. 
To train the E2E model, the training question-answer pairs $Q^{train}_{i}$ corresponding to the training document $d^{train}_i$ are converted into a flattened sentence $y_i$ the following transformation:%using a function defined as:
$$\mathcal{T}(q, a) = \text{``\texttt{question:} } \{q\}, \text{ \texttt{answer:} } \{a\} \text{''}$$
$$y_{i} = ``\{\mathcal{T}(q^{1}_i, a^{1}_i)\} \mid \{\mathcal{T}(q^{2}_i, a^{2}_i)\} \mid \ldots \text{''}, $$
where each pair is textualized with the function $\mathcal{T}(q, a)$, and the textualized QA pairs are concatenated using the separator |. The E2E QA generation function $\mathcal{G}_{qa}$ is then obtained by maximizing the  conditional log-likelihood:
$$ \arg \max_{y} P_{qa}(y \mid d).$$
% \todo{Can you add a formal definition of E2E here?}

The second QA generation method is the \textbf{prompt} approach, in which the QA pair generator $\mathcal{G}_{qa}$ is an instruction-tuned LM capable of reasoning over the input prompt \( I \). In this approach, the QA pairs are generated as follows: $\mathcal{G}_{qa}(d_i) = LM(d_i, I).$ We utilize Zephyr~\citep{Zephyr23Tunstall} with Chain of Thought (CoT)~\cite{cot22Wei} prompting for QA generation, as demonstrated in Figure~\ref{fig:qa_gen_input_generation}.
% \( QA_i = LM_{qag}(d_i, I) \). 

\subsection{Knowledge Injection with RAG}~\label{sec:know_inj_rag}

The non-parametric knowledge injection is performed using RAG, which consists of  two components: the \textit{Retriever} and the \textit{Generator}~\cite{Noise2024Florin, Reliable24Asai}.
% We assess both non-parametric and parametric knowledge injection using RAG and FT.

% Having established the corpus, we explain the RAG system, which contains two components: the \textit{Retriever} and the \textit{Generator}~\cite{Noise2024Florin, Reliable24Asai}.

\noindent
\textbf{Retriever.}
% Detailed explaination for search index: \footnote{In term-based retrieval systems such as BM25, which count the occurrences of words in documents in the corpus, the index \( I \) is a weighted bag-of-words vector. In more recent trainable neural retrieval systems, such as DPR, the index is a collection of float embeddings encoded by an encoder LM~\cite{Reliable24Asai}.} 
% ============
% Subsequently, the LM, whether the fine-tuned version \( LM_f \) or the pre-trained version \( LM_p \), uses both the original query \( q \) and the retrieved documents \( \{d_k\}_{k=0}^{K} \) to predict the output.
% ============
% BM25 calculates a score for the document-query pair based on the statistics of overlapping words:
% \[ s_{lex}(q, d) = BM25(q, d) = \sum_{t \in q \cap d} rsj_t \cdot \frac{tf_{t, d}}{tf_{t, d} + k_1 \left( (1 - b) + b \frac{|d|}{l} \right)} \]
% where \( t \) is a term, \( tf_{t,d} \) is the frequency of \( t \) in document \( d \), \( rsj_t \) is the Robertson-Spärck Jones weight~\cite{} of \( t \), and \( l \) is the average document length. The parameters \( k_1 \) and \( b \) are predefined constants~\cite{}.
The first key component in a RAG system is a retriever $R$, which builds an index for a document corpus $D$. During inference, given an input sequence $q$, the retriever identifies and ranks relevant documents $ D_q = R(D, q)$.  
In our retrieval process, we employ both sparse and dense retrievers. 
We utilize BM25~\cite{Robertson94bm25} as a sparse retriever due to its popularity and effectiveness. 
For dense retrievers, we employ DPR~\citep{Dense20Karpukhin} and Contriever~\citep{Unsupervised22Izacard} methods. 
Both models convert textual data into vector representations using a transformer network. The similarity between the query $q$ and document $d$ is defined as $S(q, d) = \vec{q} \cdot \vec{d}$, which computes the dot product between embedding vectors $\vec{q}$ and $\vec{d}$. DPR employs two independent BERT models, trained discriminatively using query-documents pairs with negative samples from BM25. Contriever, on the other hand, is trained using a shared BERT model for query and document encoding, optimized using a contrastive loss. 
We also employ a two-step retrieval pipeline, which includes first-stage retrieval using BM25 and reranking using DPR~\cite{Askari23Test, Lin2021Pretrained, Generate24Abbasiantaeb}. 

\noindent
\textbf{Generator.}
% === Describle Generator component
The second step involves a generator component responsible for synthesizing an answer, typically implemented via LMs. Generative LMs operate by predicting the probability distribution of the next token, given the previous tokens. 
In RAG, the generative LM takes a query $q$ and top-$K$ ranked documents from $D_q$, denoted as $D_q^K = [d_1, ..., d_K]$, and generates a response by sequentially predicting the next token. Our RAG prompt prepends the documents before the query, following~\cite{When23Mallen, Noise2024Florin}.
% For a given sequence of words \(w_1, w_2, \ldots, w_n\), a generative LM aims to maximize the likelihood of this sequence, expressed using the chain rule of probability:
% \[ P(w_1, w_2, \ldots, w_n) = \prod_{i=1}^{N} P(w_i|w_1, w_2, \ldots, w_{i-1}) \]
% where \(P(w_i|w_1, w_2, \ldots, w_{i-1})\) is the conditional probability of the word \(w_i\) given the preceding sequence of words \(w_1, w_2, \ldots, w_{i-1}\). % In RAG, the generative LM takes a query \(q\) and the retrieved documents \(\mathcal{D}_r\) as input and generates a response by sequentially predicting the next token in the sequence. More formally,
% \[ P_{rag}(y|q) \approx \prod_{i} \sum_{d \in \mathcal{D}_r} p_\eta(d|q)p_\theta(y_i|q, d, y_{1:i-1}), \]
% where \(p_\eta(d|q)\) is the retrieval component that provides a (truncated) probability distribution for the top-scoring documents, and \(p_\theta(y_i|q, d, y_{1:i-1})\) is a probability distribution parameterized by \(\theta\) that generates the current token based on the previously generated tokens, the query, and the retrieved document.

In this paper, we define and assess four distinct configurations of injecting knowledge with fine tuning and RAG: (1) \textit{-FT-RAG}: the vanilla LM without retrieved documents, (2) \textit{-FT+RAG}: the vanilla LM with retrieved documents, (3) \textit{+FT-RAG}: the fine-tuned LM without retrieved documents, (4) \textit{+FT+RAG}: the fine-tuned LM with retrieved documents.
% based on whether the retrieved documents are added to the input prompt and whether the pre-trained LM or the fine-tuned version is used:
% \begin{enumerate}
%     \item The pre-trained LM without retrieved documents (\textit{-FT-RAG})
%     \item The pre-trained LM with retrieved documents (\textit{-FT+RAG})
%     \item The fine-tuned LM without retrieved documents (\textit{+FT-RAG})
%     \item The fine-tuned LM with retrieved documents (\textit{+FT+RAG})
% \end{enumerate}

\begin{figure}[t]
  \centering
  \includegraphics[width=0.46\textwidth]{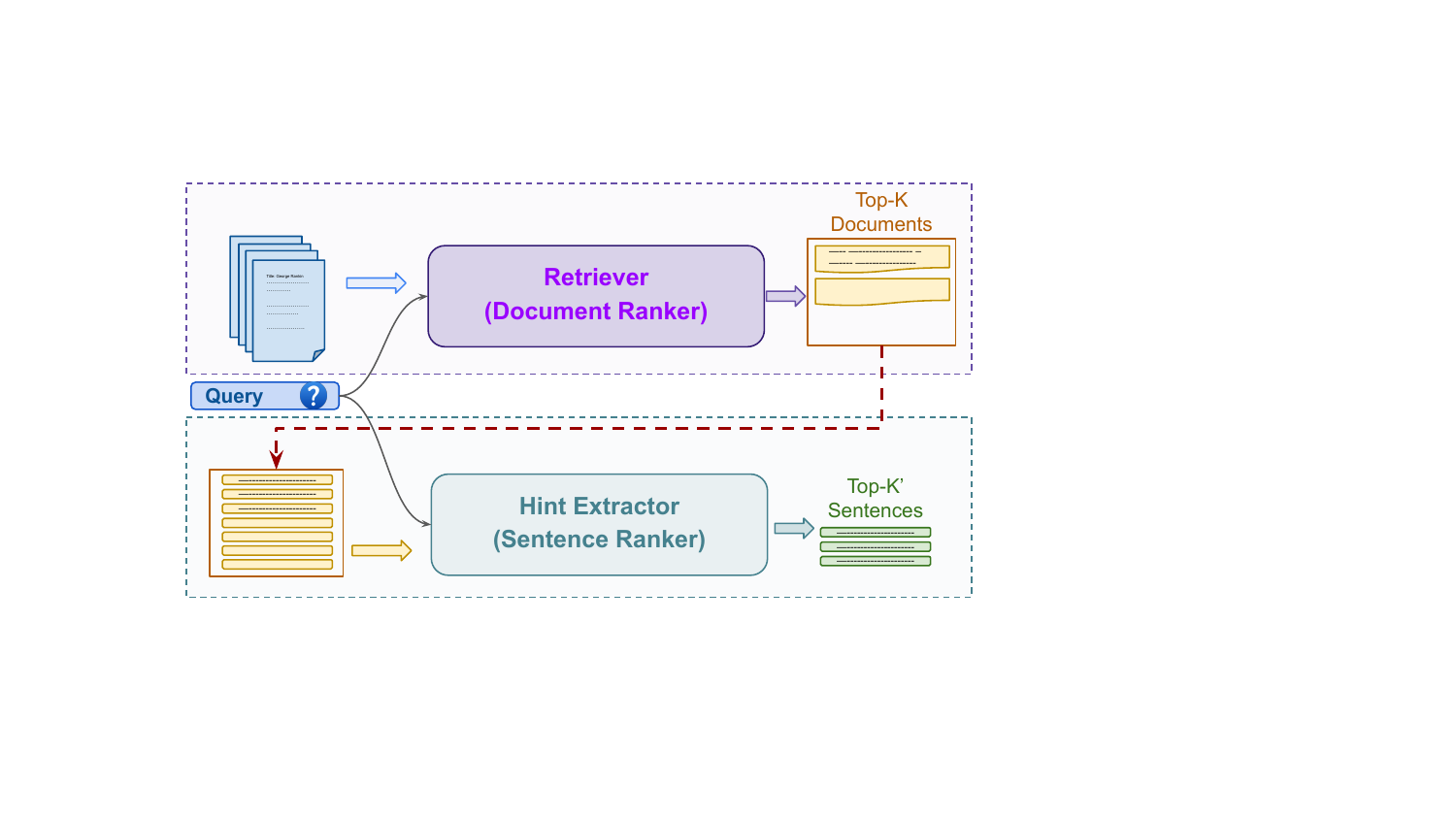}
  \shrink
  \caption{\label{fig:highlight_model} Our proposed Stimulus RAG method. The \textit{Hint Extractor}  identifies the most relevant sentence from top-K documents ranked by the retriever. This sentence is then added to the beginning of the input prompt.}% to guide the LMs in answering the query.}
\shrink

\end{figure}

% ========== Highlighting method ==================
\subsection{Stimulus RAG}

% === Problem definition
While the generic retrieve-then-generate framework of RAG is effective in answering factual knowledge~\cite{When23Mallen}, it sometimes struggles to accurately respond to factual questions that are not memorized in LMs parameters, even when the ground truth document is included in the prompt. 
We hypothesize that by employing a more advanced RAG approach we can achieve or even surpass the benefits of the combined RAG and FT (\textit{+FT+RAG}) setup. This would allow us to bypass the computationally expensive process of knowledge injection with FT, which involves resource-intensive processes of QA generation and FT.

% Knowledge injection with fine tuning  is more resource-intensive compared to the Retrieval-Augmented Generation (RAG), as it involves the two steps of generating QA pairs and the fine-tuning. Therefore, we pose the question, \textit{``Can the fine tuning step be bypassed with a more advanced RAG system?''}~\ref{rq:3}. 
% In other words, can we achieve the accuracy of \textit{+FT+RAG} by enhancing our RAG system in \textit{-FT+RAG}?
% To address this, we first need to understand, "What do LMs learn during fine-tuning?" To explore this, we analyzed the outputs of the LMs. Our findings indicate that LMs acquire high-level information, such as the task type or the nature of the relationships being questioned. Additionally, we observed that LMs sometimes understand the type of entities to be selected but cannot identify the exact answer.
% To mitigate these issues, we propose a highlighting method to help LMs find the correct answer. The idea is that if we can identify the answer-containing part of a document, we can guide the LM to focus more on the highlighted section. In this approach, we will pose our questions to the LM, directing it to find the answer from the given documents with increased emphasis on the highlighted part. 

We introduce \emph{Stimulus RAG}, a RAG approach that guides LMs to generate responses using a hint provided in the RAG prompt. Stimulus RAG comprises three steps: \textit{Retriever}, \textit{Hint extractor}, and \textit{Generator}. The Retriever and Generator are as defined in Section~\ref{sec:know_inj_rag}. The \textit{\textbf{Hint extractor}} provide hint text that guides LM to generate accurate responses and works as follows: The top-$K$ ranked documents $D_q^K = [d_1, ..., d_K]$ ($d_K$ denoting the retrieved document at rank $K$) from the retriever step are split into sentences, denoted as $S_q=\{s_j\}_{j=0}^{N}$, where $N$ is the total number of sentences. These sentences are then ranked using a retrieval model $R'(S_q, q)$.  For the sake of simplicity, we use the same retrieval model as that used in the retriever step, hence $R=R'$.
% The retrieval model $R'$ can be similar or different from the model in the retriever step $R$
% , we then rank $\{s_n\}_{n=0}^{N}$ for the query $q$.
The top-ranked sentence $s_q^1$ is identified as the most relevant sentence for the given question and serves as a hint. 
The sentence $s_q^1$ can be directly used in the prompt, which we refer to as the SRAG(S) approach. Alternatively, the document containing the sentence $s_q^1$ can be considered as a hint, which we refer to as the SRAG(D) approach.
The extracted hint is placed at the top of the input prompt provided to the \textit{\textbf{Generator}} component. This implies that the hint text a repetition of presumably the most relevant sentence/document of in the prompt. It has been shown that the beginning of the input prompt receives more attention from LMs~\cite{Lost2023Nelson}. Our method is illustrated in Figure~\ref{fig:highlight_model}.

\begin{table}[t]
% \small
\centering
\caption{Statistics of the factual knowledge-based datasets. 
% These datasets were selected because their queries focus on factual knowledge and are sourced from Wikipedia.
}
\shrink
\label{tb:dataset_statistic}
\begin{tabular}{
% @{~}l@{~~~~}|c@{~~~}c@{~~~}c@{~}
l|lcl
}
\hline
\textbf{Dataset} &
\textbf{\# QA} &
\textbf{\# Rel. Type} &
\textbf{Question form} \\
\hline
\textsc{PopQA} & 14K & 16 & Template \\
\textsc{WiTQA} & 14K & 32 & Model-assisted \\
EQ & 17.3K & 24 & Template \\
\hline
\end{tabular}
\end{table}

\begin{figure}[t]
    \centering
    \includegraphics[width=0.46\textwidth]{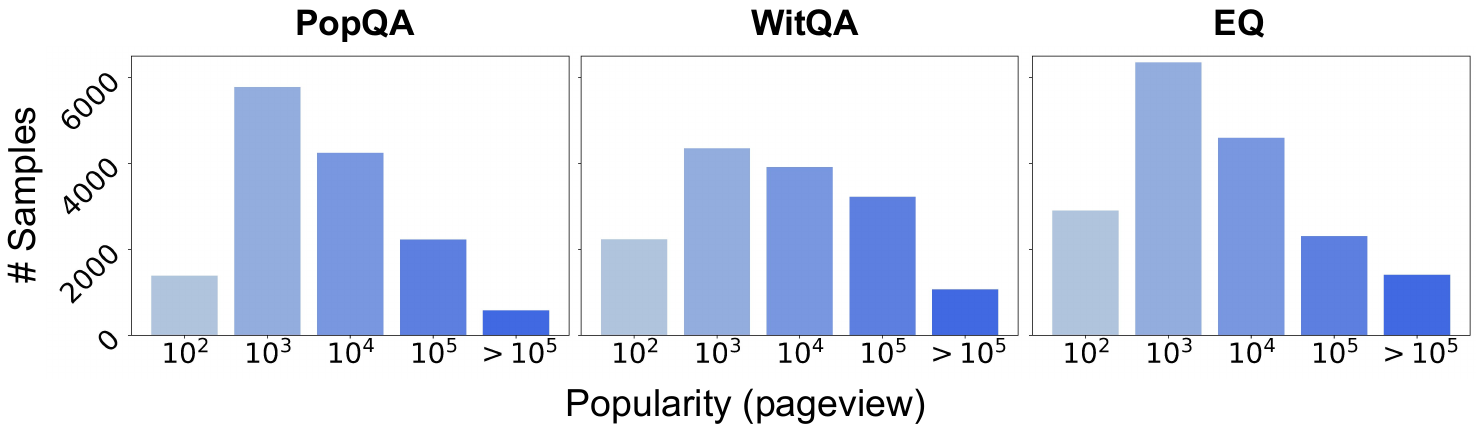}
    \shrink
    \caption{Distribution of sample counts across popularity buckets, defined by $log_{10}(\text{pageviews})$ for \textsc{PopQA} and \textsc{WitQA} and $log_{2}(\text{pageviews})$ for \textsc{EQ}.
    % For example, the leftmost bin includes entities with fewer than $10^2$ pageviews, while the rightmost bin encompasses entities with over $10^5$ pageviews for \textsc{PopQA} and \textsc{WitQA}.
    }
    \label{fig:bk_dist}
    % \shrink
\end{figure}

\section{Experimental Setup}~\label{sec:experimental_setup}

% === Table 1: version 2 with 2 columns
\renewcommand{\arraystretch}{1.25}
\begin{table}[t]
\small
\centering
\caption{Accuracy of vanilla and fine-tuned LMs, both with and without RAG. The RAG results are based on ideal retrieval. Statistically significant differences in the \textit{PEFT-Prompt} rows are compared with other rows. Superscripts (a), (b), (c), and (d) indicate statistically significant differences (better or worse) compared to vanilla LM, \textit{PEFT-E2E}, \textit{Full-E2E}, and \textit{Full-Prompt}, respectively, determined by the Wilcoxon test (p-value < $0.01$).
}
\shrink
\label{tb:ft_qa_results}
\begin{tabular}{@{~}
ll|ll|ll
% l@{~~~~}c@{~~~} | c@{~~~}c@{~~~} | c@{~~~}c@{~}
}
\hline
\textbf{~~} & \textbf{~~} & \multicolumn{2}{l}{\textbf{\textsc{PopQA}}} & \multicolumn{2}{l}{\textbf{\textsc{EQ}}} \\
\hline

% \textbf{~~} & \textbf{~~} &
% \textbf{-FT/} & \textbf{-FT/} & \textbf{+FT/} & \textbf{+FT/} &
% \textbf{-FT/} & \textbf{-FT/} & \textbf{+FT/} & \textbf{+FT/} \\
\textbf{FT} & \textbf{QA} & \textbf{+FT-RAG} & \textbf{+FT+RAG} & \textbf{+FT-RAG} & \textbf{+FT+RAG} \\
\hline

% \multicolumn{6}{l}{\textbf{FlanT5-small}} \\ \hline
\multicolumn{2}{l|}{\textbf{FlanT5-small}} & 2.69 & 47.46 & 2.84 & 27.36 \\ \hline

% \multirow{2}{*}{PEFT}
PEFT & E2E    & 6.06 & 48.40 & 7.86 & 33.76 \\ 
PEFT & Prompt & 7.01\textsuperscript{(a,c,d)} & 61.39\textsuperscript{(a,b,c,d)} & 9.39\textsuperscript{(a,b,c,d)} & 42.68\textsuperscript{(a,b,c,d)} \\
Full & E2E    & 5.19 & 12.63 & 10.98 & 21.27 \\ % [] [v109]
Full & Prompt & 8.55 & 46.88 & 15.52 & 38.38 \\ % [] [v108]
\hline

% \multicolumn{6}{l}{\textbf{FlanT5-base}} \\ \hline
\multicolumn{2}{l|}{\textbf{FlanT5-base}} & 6.01 & 73.08 & 6.07 & 53.92 \\ \hline
PEFT & E2E    & 7.53 & 70.34 & 10.98 & 51.30 \\ % [] [v114]
PEFT& Prompt & 9.11\textsuperscript{(a,b,c)} & 71.34\textsuperscript{(a,b,c,d)} & 12.98\textsuperscript{(a,b,c,d)} & 57.63\textsuperscript{(a,b,c,d)} \\
Full & E2E    & 7.42 & 44.76 & 10.91 & 31.22 \\ % [v111] [v110]
Full & Prompt & 10.06 & 51.80 & 17.36 & 54.07 \\ % [] [115]
\hline

% \multicolumn{6}{l}{\textbf{FlanT5-large}} \\ \hline
\multicolumn{2}{l|}{\textbf{FlanT5-large}} & 8.44 & 68.56 & 16.94 & 52.64 \\ \hline

PEFT & E2E    & 8.69 & 67.47 & 15.33 & 53.25 \\ % [v113] [v114]
PEFT & Prompt & 11.24\textsuperscript{(a,b,d)} & 71.27\textsuperscript{(a,b,c,d)} & 18.17\textsuperscript{(a,b,c)} & 60.08\textsuperscript{(a,b,c,d)} \\
Full& E2E    & 11.75 & 27.31 & 14.79 & 23.17 \\ % [v122] [v123]
Full & Prompt & 13.60 & 68.18 & 18.22 & 57.37 \\ % [v120] [v121]
\hline

% \multicolumn{6}{l}{\textbf{StableLM2}} \\ \hline
\multicolumn{2}{l|}{\textbf{StableLM2}} & 17.01 & 76.14 & 17.92 & 60.72 \\ \hline

PEFT & E2E    & 16.39 & 74.82 & 23.62 & 58.34 \\  % [v117] [v118]
PEFT & Prompt & 21.75\textsuperscript{(a,b,c,d)} & 82.09\textsuperscript{(a,b,c,d)} & 27.23\textsuperscript{(a,b,c,d)} & 68.82\textsuperscript{(a,b,c,d)} \\
Full & E2E    & 14.23 & 5.87 & 13.22 & 2.46 \\ % [v126] [v127]
Full & Prompt & 25.73 & 32.50 & 23.22 & 30.07
\\ % [v124] [v125 --]
\hline

\end{tabular}
% \shrink
% \vskip -0.1in
\end{table}

% === Table 2: 
\renewcommand{\arraystretch}{1.25}
\begin{table*}[t]
    % \small
    \centering
    \caption{Accuracy of vanilla and fine-tuned LMs, both with and without RAG. The RAG results are based on Ideal retrieval. The PEFT method is used for fine-tuning, and prompting is utilized for QA generation. The best results in each column are shown in \textbf{bold}, and the best results in each row are \underline{underlined}. Statistically significant differences in the \textit{+FT+RAG} columns are compared with the \textit{-FT+RAG} columns. Superscript (a) indicates statistically significant differences (better or worse) as determined by the Wilcoxon test (p-value < $0.01$).}
    \shrink
    \label{tb:all_results}
    \begin{tabular}{@{~}l@{~~~~}| l@{~~~}
    | llll
    | llll
    | lllll@{~}}
    % | l@{~~~~~~}l@{~~~~~~}l@{~~~~~~}l@{~~~~~~}
    % | l@{~~~~~~}l@{~~~~~~}l@{~~~~~~}l@{~}}
    \hline
    \textbf{Dataset} &  &
    \multicolumn{4}{c}{\textbf{\textsc{PopQA}}} &
    \multicolumn{4}{c}{\textbf{\textsc{WiTQA}}} &
    \multicolumn{4}{c}{\textbf{EQ}}
    \\ \hline
    \textbf{Model}   & \textbf{\#P} &
    \textbf{-FT} & \textbf{+FT} & \textbf{-FT} & \textbf{+FT} &
    \textbf{-FT} & \textbf{+FT} & \textbf{-FT} & \textbf{+FT} &
    \textbf{-FT} & \textbf{+FT} & \textbf{-FT} & \textbf{+FT}
    \\
    \textbf{~~}& &
    \textbf{-RAG} & \textbf{-RAG} & \textbf{+RAG} & \textbf{+RAG} &
    \textbf{-RAG} & \textbf{-RAG} & \textbf{+RAG} & \textbf{+RAG} &
    \textbf{-RAG} & \textbf{-RAG} & \textbf{+RAG} & \textbf{+RAG}
    \\ \hline
    FlanT5-small & 80m &
    % 2.69  & 6.93  & 47.48 & \textbf{59.36} & []% old 
    % [v1: 2.69][v2: 2.09] & [v1: 7.26][v2: 7.19] & [v1: 47.46][v2: 67.65] & [v1: 61.39][59.36][v2: 60.72]
    2.69  & 7.26  & 47.46 & \underline{61.39}\textsuperscript{(a)} &
    8.76  & 18.30 & 40.76 & \underline{59.47}\textsuperscript{(a)} & % [64.91] [59.02]
    2.84  & 9.39  & 27.36 & \underline{42.68}\textsuperscript{(a)} \\% [49.46] [44.12]
    FlanT5-base  & 250m &
    6.01  & 9.11  & \underline{73.08} & 71.34\textsuperscript{(a)} &
    16.52 & 23.32 & 73.35 & \underline{74.34} &
    6.07  & 12.98 & 53.92 & \underline{57.63}\textsuperscript{(a)} \\
    FlanT5-large & 780m & 
    8.44  & 11.24 & 68.56 & \underline{71.27}\textsuperscript{(a)} &
    24.52 & 28.85 & 74.37 & \underline{77.24}\textsuperscript{(a)} &
    16.94 & 18.17 & 52.64 & \underline{60.08}\textsuperscript{(a)} \\  
    Tiny-llama   & 1.1B &
    17.50 & 18.32 & 74.39 & \underline{74.87} &
    45.12 & 47.65 & 78.84 & \underline{80.60}\textsuperscript{(a)} &
    21.12 & 24.57 & 61.03 & \underline{61.61} \\
    StableLM2    & 1.6B &
    17.01 & 21.75 & 76.14 & \underline{\textbf{82.09}}\textsuperscript{(a)} &
    42.18 & 51.66 & 81.08 & \underline{{86.19}}\textsuperscript{(a)} &
    17.92 & 27.23 & 60.72 & \underline{\textbf{68.82}}\textsuperscript{(a)} \\
    MiniCPM      & 2B & 
    14.16 & 15.47 & 69.44 & \underline{75.86}\textsuperscript{(a)} &
    37.61 & 44.94 & 73.17 & \underline{80.45}\textsuperscript{(a)} &  
    15.31 & 22.92 & 54.63 & \underline{62.67}\textsuperscript{(a)} \\
    FlanT5-xl    & 3B &
    12.24 & 13.25 & 73.31 & \underline{74.71}\textsuperscript{(a)} & % [5952736] 
    31.24 & 36.38 & 76.97 & \underline{79.67}\textsuperscript{(a)} & % [5958030] 
    15.98 & 20.42 & 59.07 & \underline{62.70}\textsuperscript{(a)} \\% [5958099] 
    Mistral      & 7B &
    21.47 & 30.70 & \underline{80.25} & 78.44\textsuperscript{(a)} &
    51.36 & 58.29 & \underline{86.05} & 83.90\textsuperscript{(a)} &
    25.78 & 34.01 & \underline{68.60} & 64.96\textsuperscript{(a)} \\
    Zephyr       & 7B &
    {28.23} & \textbf{35.48} & \underline{78.65} & 78.20 &
    {58.33} & \textbf{63.74} & 83.83 & \underline{\textbf{86.89}}\textsuperscript{(a)} &
    {29.09} & \textbf{38.52} & 62.45 & \underline{67.89}\textsuperscript{(a)} \\
    Llama2-chat  & 7B &
    26.09 & 27.71 & \underline{81.15} & 80.15 &
    53.88 & 56.38 & \underline{86.50} & 84.71\textsuperscript{(a)} &
    27.13 & 32.84 & \underline{68.29} & 67.11\textsuperscript{(a)} \\
    Llama3-chat   & 8B &
    \textbf{32.52} & 32.75 & \textbf{81.29} & \underline{81.54} &
    \textbf{61.88} & 61.58 & \underline{\textbf{86.86}} & 85.71 &
    \textbf{35.07} & 37.95 & \underline{\textbf{68.67}} & 68.64 \\
    FlanT5-xxl   & 11.3B &
    11.26 & 15.94 & \underline{75.19} & 74.98 &
    30.40 & 42.89 & 78.19 & \underline{{81.44}}\textsuperscript{(a)} &
    12.48 & 23.16 & 59.67 & \underline{{62.80}}\textsuperscript{(a)} \\
    \hline
    \end{tabular}
\end{table*}

\noindent
\textbf{Datasets.}
We conduct our experiments on three datasets focused on factual knowledge: \textsc{PopQA}~\cite{When23Mallen}, \textsc{WitQA}~\cite{maekawa24witqa}, and EntityQuestion (EQ)~\cite{Simple21Sciavolino}, all of which include long-tail entities; see Table~\ref{tb:dataset_statistic} for statistics. \textsc{PopQA} is an open-domain QA dataset about long-tail entiteis, constructed from 16 diverse relationship types in Wikidata. EQ is another popular open-domain QA dataset that covers a long-tail entity distribution, using Wikipedia hyperlink counts as a proxy for entity frequency and sampling knowledge triples from Wikidata based on these frequency distributions. Since EQ does not provide Wikidata IDs for each entity, we use only about 80\% of the questions, where the mention of the subject entity has a unique match with a Wikidata entity.
\textsc{WitQA} is another entity-centric dataset that defines a different proxy for popularity. They argue that the popularity metric should be based on the occurrence of both the subject entity and the relation (unlike the subject-based popularity in \textsc{PopQA} and EQ). Therefore, they define the S-R count, which is the co-occurrence of the subject entity and relation predicate. However, they report pageviews in their dataset, and we use the pageview-based popularity to enable comparison with other datasets.
To analyze performance with respect to popularity, we divide the entities into five buckets based on their popularity levels; see Figure~\ref{fig:bk_dist}.

\medskip \noindent
\textbf{Evaluation Metric.}
Following previous studies \cite{When23Mallen, maekawa24witqa, Simple21Sciavolino}, we report on the accuracy metric, where a prediction is considered correct if one of the ground truth responses matches a substring of the predicted response.
While widely used~\cite{Large23Kandpal, Lost2023Nelson, Noise2024Florin}, this metric is not without issues.
A principal problem arises in determining response correctness, particularly in cases involving date representations or varying phrasings that convey identical meanings~\cite{Noise2024Florin}. For example, comparing the predicted response ``Nathanson'' with the ground truth ``Jeff Nathanson,'' the prediction is considered incorrect.
Another observed problem is that when the model generates multiple entity names, among them the  ground truth entity,  the response is incorrectly considered as correct.
Recognizing these limitations, we acknowledge the necessity for a more advanced analysis of answer variations, which we leave for future research.
%a prediction is considered correct if it contains a substring that exactly matches any of the provided gold answers. 
% Although this metric is used in other relevant work~\cite{Large23Kandpal, Lost2023Nelson, Noise2024Florin}, it is not without issues. % in response to a query where the correct answer is deemed incorrect “Jeff Nathanson,” the response would be deemed incorrect under our current evaluation schema. 

\begin{table*}[t]

\centering
\caption{Performance of retrievers. The relevant document for each question is assumed to be the first paragraph of the corresponding Wikipedia page. Statistically significant differences in the \textit{DPR} row are compared with other rows. Superscripts (a), (b), and (c) indicate statistically significant differences (better or worse) compared to \textit{BM25}, \textit{Contriever}, and \textit{BM25+DPR}, respectively, as determined by a t-test (p-value < $0.01$).}
\shrink
\label{tb:ret_res}
\begin{tabular}{l| lll | lll | lll}
\hline
\textbf{Dataset} &
\multicolumn{3}{c}{\textbf{\textsc{PopQA}}} &
\multicolumn{3}{c}{\textbf{\textsc{WiTQA}}} &
\multicolumn{3}{c}{\textbf{EQ}}
\\ \hline
\textbf{Model}&
\textbf{Rec@1} & \textbf{Rec@3} & \textbf{Rec@5} &
\textbf{Rec@1} & \textbf{Rec@3} & \textbf{Rec@5} &
\textbf{Rec@1} & \textbf{Rec@3} & \textbf{Rec@5} \\

\hline
BM25       &
40.13 & 62.92 & 71.30 &
56.40 & 81.88 & 88.87 &
64.01 & 89.00 & 93.71 \\
Contriever &
42.90 & 70.84 & 80.17 &
53.23 & 84.25 & 91.52 &
63.10 & 91.93 & 96.30 \\
BM25+DPR   &
59.35 & 80.85 & 87.55 &
72.01 & 90.17 & 93.72 &
79.10 & 95.09 & 96.11 \\
DPR        &
59.40\textsuperscript{(a, b)} &
80.85\textsuperscript{(a, b)} &
87.57\textsuperscript{(a, b)} &
71.73\textsuperscript{(a, b)} &
90.05\textsuperscript{(a, b)} &
93.62\textsuperscript{(a, b)} &
79.01\textsuperscript{(a, b)} &
94.06\textsuperscript{(a, b)} &
96.10\textsuperscript{(a)} \\
\hline
\end{tabular}
\end{table*}

\renewcommand{\arraystretch}{1.25}
\begin{table*}[t]
    \centering
    \small
    \caption{
    Overall accuracy of answer generation using different retrievers before and after FT. The performance of LMs in answering questions correlates with the effectiveness of the retrieval models. Statistically significant differences in the \textit{DPR} rows are compared with other rows. Superscripts (a), (b), and (c) indicate statistically significant differences (better or worse) compared to \textit{BM25}, \textit{Contriever}, and \textit{BM25+DPR}, respectively, as determined by the Wilcoxon test (p-value < $0.01$).
    }
    \shrink
    \label{tb:answer_generator_retrievers}
    \begin{tabular}{ 
    l|c|cccc|cccc|cccc
    % @{~}l@{~~} 
    % | c@{~~~}
    % | c@{~~~}c@{~~~}c@{~~~}c@{~~~}
    % | c@{~~~}c@{~~~}c@{~~~}c@{~~~}
    % | c@{~~~}c@{~~~}c@{~~~}c@{~}
    }
    
    \hline
    & & 
    \multicolumn{4}{c}{\textbf{\textsc{PopQA}}} &
    \multicolumn{4}{c}{\textbf{\textsc{WiTQA}}} &
    \multicolumn{4}{c}{\textbf{EQ}}
    \\ \hline
    \textbf{Model} & &
    % \textbf{\rotatebox[origin=c]{90}{BM25}} &
    % \textbf{\rotatebox[origin=c]{90}{Contriever}} &
    % \textbf{\rotatebox[origin=c]{90}{BM25+DPR}} &
    % \textbf{\rotatebox[origin=c]{90}{DPR}} &
    \textbf{BM25} &
    \textbf{\makecell{Contr-\\iever}} &
    \textbf{\makecell{BM25+\\DPR}} &
    \textbf{DPR} &
    
    \textbf{BM25} &
    \textbf{\makecell{Contr-\\iever}} &
    \textbf{\makecell{BM25+\\DPR}} &
    \textbf{DPR} &
    
    \textbf{BM25} &
    \textbf{\makecell{Contr-\\iever}} &
    \textbf{\makecell{BM25+\\DPR}} &
    \textbf{DPR} \\
    \hline

    FlanT5-base & -FT & 
    41.40 & 44.48 & 54.21 & 53.36\textsuperscript{(a, b)} &
    60.25 & 58.06 & 66.62 & 66.51\textsuperscript{(a, b)} &
    43.90 & 43.01 & 49.62 & 49.54\textsuperscript{(a, b)} \\
    FlanT5-base & +FT & 
    39.99 & 43.24 & 52.19 & 52.21\textsuperscript{(a, b)} &
    61.06 & 59.05 & 66.89 & 66.83\textsuperscript{(a, b)} &
    45.35 & 45.24 & 50.45 & 50.39\textsuperscript{(a, b)} \\

    StableLM2
    & -FT & 
    46.05 & 47.83 & 55.49 & 55.74\textsuperscript{(a, b)} &
    72.52 & 69.33 & 74.53 & 74.47\textsuperscript{(a, b)} &
    51.71 & 49.25 & 54.60 & 54.51\textsuperscript{(a, b)} \\
    StableLM2 & +FT & 
    49.92 & 53.63 & 60.97 & 61.44\textsuperscript{(a, b)} &
    77.67 & 76.65 & 80.48 & 80.45\textsuperscript{(a, b)} &
    59.99 & 59.24 & 63.67 & 61.91\textsuperscript{(a, b)} \\

    Mistral
    & -FT & 
    49.58 & 53.10 & 60.14 & 60.09\textsuperscript{(a, b)} &
    78.33 & 77.35 & 81.54 & 81.49\textsuperscript{(a, b)} &
    59.09 & 58.03 & 63.13 & 63.06\textsuperscript{(a, b)} \\
    Mistral & +FT & 
    50.53 & 54.08 & 61.38 & 61.23\textsuperscript{(a, b)} &
    76.82 & 75.60 & 79.35 & 79.22\textsuperscript{(a, b)} &
    56.74 & 55.19 & 61.43 & 61.28\textsuperscript{(a, b)} \\

    Llama3 & -FT & 
    51.30 & 56.49 & 61.57 & 61.46\textsuperscript{(a, b)} &
    78.51 & 79.00 & 81.62 & 81.73\textsuperscript{(a, b)} &
    59.31 & 59.26 & 63.37 & 63.27\textsuperscript{(a, b)} \\
    Llama3& +FT & 
    53.57 & 56.90 & 62.05 & 62.10\textsuperscript{(a, b)} &
    79.96 & 79.59 & 81.72 & 81.77\textsuperscript{(a, b)} &
    61.86 & 61.05 & 64.07 & 64.00\textsuperscript{(a, b)} \\

    \hline
    \end{tabular}
    % \shrink
% \vskip -0.1in
\end{table*}

\medskip \noindent
\textbf{Language Models.}
We use several LMs, focusing on two main features: the backbone architecture (i.e., decoder-only and encoder-decoder) and model size, which ranges from 80 million to over 11 billion parameters. For the encoder-decoder models, we utilize five versions of the FlanT5 family~\cite{Scaling22Chung}, spanning from small to XXL. For the decoder-only models, we employ smaller LMs such as Tiny-llama~\cite{tinyllama2024Peiyuan}, StableLM2~\cite{Stablelm224Bellagente}, and MiniCPM~\cite{MiniCPM24Hu}, which range from 1 to 2 billion parameters. Additionally, we incorporate larger LMs like Mistral~\cite{Mistral23Jiang}, Zephyr~\cite{Zephyr23Tunstall}, LLama2~\cite{llama223Touvron}, and LLama3,\footnote{\url{https://ai.meta.com/blog/meta-llama-3}} which range from 7 to 8 billion parameters.
As for instructions, we apply zero-shot prompting for generative prediction using a straightforward template.
For non retrieved-augmented input, we use template:
\noindent
\verb|"Question: <question>"|, and for retrieved-augmented input, we use template:
\verb|"Context:| \verb|<context> Question: <question>"|.

\medskip \noindent
\textbf{Fine Tuning.}
We generate training data for the FT approach using two distinct data augmentation methods. 
% The first method is the End-to-End (E2E) approach~\citep{Empirical2023ushio}, which employs a model specifically trained for paragraph-level QA generation, utilizing T5-large~\citep{Exploring20Raffel}. This method is referred to as \textbf{E2E} in our paper. Additionally, we explore generating synthetic training data by prompting a LM, specifically using Zephyr~\citep{Zephyr23Tunstall} for QA generation (Figure~\ref{fig:qa_gen_input_generation}), referred to as the \textbf{Prompt} method in our discussion.
To ensure a fair comparison between FT and RAG with an ideal retriever, we generate QAs exclusively using the summary sections of Wikipedia pages.
After generating QA pairs, we proceed to fine-tune LMs using two approaches: full parameter tuning (\textbf{Full}) and Parameter Efficient Fine-Tuning (\textbf{PEFT}). Within the range of PEFT techniques~\citep{BitFit22Zaken, Few22Haokun, fine23Ma}, we utilize QLoRA~\citep{QLoRA23Dettmers}, chosen for its broad acceptance in the field~\citep{Challenges23Kaddour, Naveed23Comprehensive} and efficient use of computational resources we had at our disposal.

\medskip \noindent
\textbf{RAG.}
We utilize a variety of retrieval models to obtain relevant documents for the RAG approach, including BM25~\citep{BM2509Robertson}, Contriever~\citep{Unsupervised22Izacard}, DPR~\citep{Dense20Karpukhin}, and a two-stage re-ranker that combines BM25 with DPR, all implemented according to the BEIR benchmark~\citep{BEIR21Thakur}.
Additionally, since the selected datasets do not contain grounded document evidence, we assume that the summary section of each Wikipedia page is the answer-containing document. We define an ideal retriever model as one that returns the summary paragraph as the top-ranked document, referred to as the \textbf{Ideal} retriever throughout the paper. We acknowledge that this assumption is not entirely accurate as some answers may be found in other subsections. However, our evaluation of the downstream task (Figures \ref{fig:all_pop_flant5} and \ref{fig:ret_res_bk}) demonstrates that the Ideal retriever outperforms other retrievers. % and is not biased by popularity or relation types.

\medskip \noindent
\textbf{Stimulus RAG.}
We select a DPR retriever for the \textit{hint extractor} and set $K = 3$. We report on two variations of the SRAG approach: (i) SRAG(S), which utilizes the top-1 sentence as the hint, and (ii) SRAG(D), which inserts the entire document contacting the top-ranked sentence. 
% \todo{explain the prompt with hint}
For the instruction, we used the following template: \verb|"Context:| \verb|<hint><context> Question: <question>"|.
\section{Results}~\label{sec:results}

In the following, we evaluate fine-tuning and RAG methods on different setups and answer our two research questions listed in Section~\ref{sec:intro}.

% In the remainder of this section, we address two questions:
% \ref{rq:1} To what extent can factual knowledge be memorized by LMs, how can the memorization process be optimized considering available resources, and what factors affect this memorization?
% \ref{rq:2} How does the performance of retrieval-augmented LMs compare to non-customized and customized LMs?

\subsection{Fine Tuning vs. RAG Performance}
The first research question \emph{\textbf{(RQ1)}} involves the comparison between FT and RAG methods and factors affecting these models: (i) fine-tuning method, (ii) data augmentation method, (iii) LM type and size, and (iv) retrieval model performance.

\medskip \noindent
\textbf{Comparison of fine tuning methods.}
To study the effect of fine-tuning method, we investigate four LMs: three from the FlanT5 family (encoder-decoder models) and StableLM2 (a decoder-only model). These models are chosen as their parameters are less than 2B, allowing us to perform full fine-tuning with the resources we had at our disposal. 
% chose To address part of \emph{\textbf{(RQ1)}}, which concerns the factors affecting memorization, we explore two key aspects that influence model specialization in processing less popular knowledge. A key aspect under review is the effect of full tuning versus PEFT. We investigate four LMs: three from the FlanT5 family and StableLM2, a decoder-only model. The Ideal retriever is used for the RAG part. 
Table~\ref{tb:ft_qa_results} shows that PEFT leads to smaller gains in the \textit{+FT-RAG} setup compared to full FT in most cases, yet it significantly improves accuracy in the \textit{+FT+RAG} setup. This suggests that PEFT enables the LM to maintain its reasoning abilities based on the provided prompts. Based on this observation, we selected PEFT as the fine-tuning method for subsequent experiments.

% (see Table~\ref{tb:generated_qa_details} in Appendix~\ref{ap:ft_setup})
\medskip \noindent
\textbf{Comparison of data augmentation methods.}
The E2E method generates over 12 times more QAs than the prompting method, while the prompt-based method generates higher-quality QAs. The results in Table~\ref{tb:ft_qa_results} show that fine-tuned models trained on prompt-generated data outperform those trained on E2E-generated data. This highlights the significance of synthetic data quality over quantity. As a result, for subsequent experiments, we use QAs generated by the prompt method.

% \begin{figure}[t]
%   \centering
%   \includegraphics[width=0.46\textwidth]{src/all_llms.png}
%   \caption{\label{fig:all_pop_llms} The performance of decoder-only LMs with different combinations of FT and RAG. These LMs memorize popular knowledge very well, so using RAG only slightly improves performance. In contrast, RAG is highly effective for less popular knowledge.}
% \end{figure}

\medskip \noindent
\textbf{Comparison of FT and RAG with LMs of different type and size.}
Choosing PEFT as the FT method and the Prompt method for QA generation, we extended our experiments to 12 LMs using the ideal retriever for RAG. Table~\ref{tb:all_results} presents the results. The findings indicate that while FT enhances the answer generator's performance, it alone cannot match or approach the performance of RAG. However, combining FT with RAG yields the best results for smaller models (up to 3B parameters). Conversely, for larger models (from 7B to 11.3B parameters), combining FT with RAG degrades performance. This suggests that although FT injects knowledge into LMs (as seen when comparing \textit{-FT-RAG} with \textit{+FT-RAG}), it diminishes the reasoning abilities of larger LMs.
Interestingly, since smaller LMs' accuracy improves with FT, the best result for \textit{+FT+RAG} is achieved by StableLM2, which has only 1.6B parameters. Another observation is that decoder-only models perform better than encoder-decoder (Flan-T5) models of similar size.

\begin{figure}[t]
  \centering
  \includegraphics[width=0.48\textwidth]{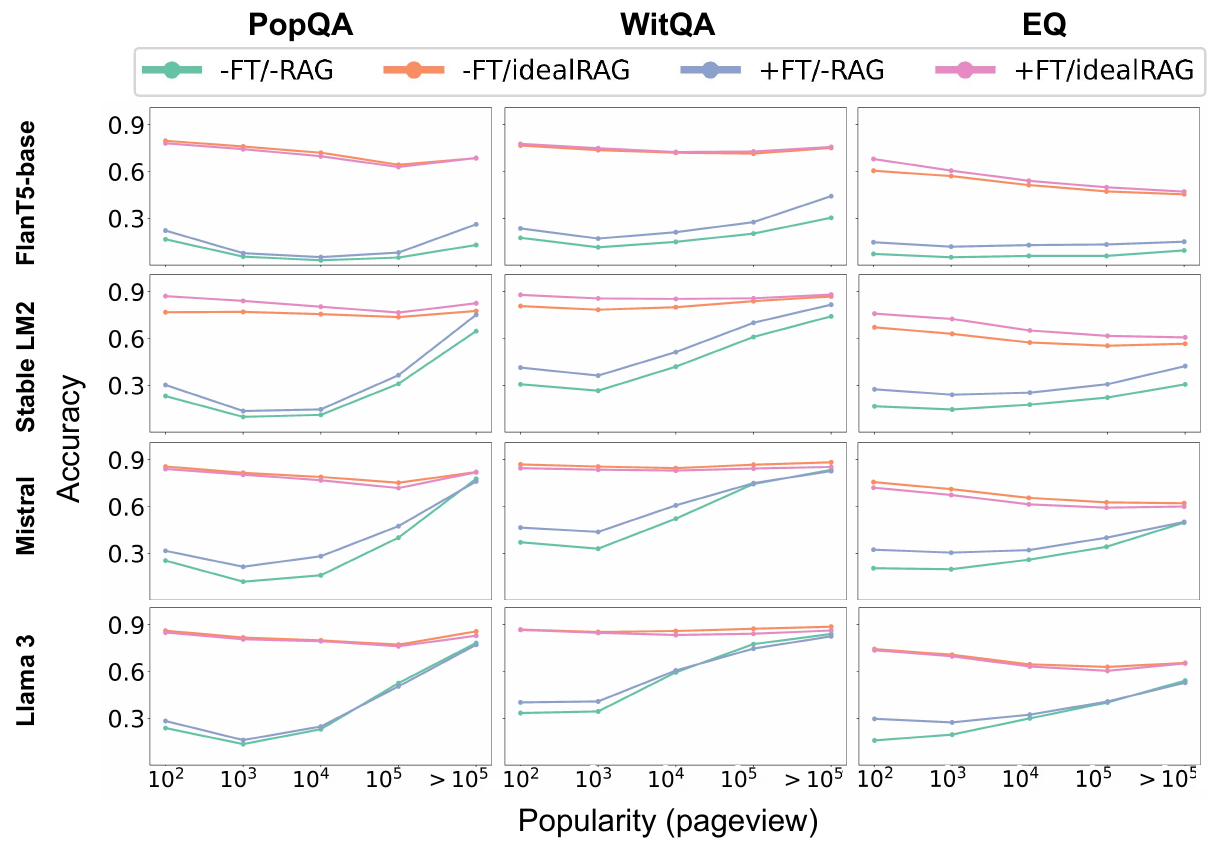}
  \shrink
  \shrink
  \caption{\label{fig:all_pop_flant5} The performance of the LMs with different combinations of FT and RAG. The \textit{+FT+RAG} setup outperforms other setups across all models and datasets.}
  \shrink
\end{figure}

% Figures \ref{fig:all_pop_flant5} and \ref{fig:all_pop_llms} show the accuracy across different popularity buckets to investigate the effectiveness of RAG and FT at various popularity levels. It is observable that RAG increases the accuracy for the least popular entities, which aligns with \citet{When23Mallen}'s findings. Moreover, these figures demonstrate that fine-tuning enhances QA accuracy across all popularity levels, with the greatest improvements observed in the most and least popular buckets for certain LMs, such as FlanT5-small, FlanT5-base, and FlanT5-XXL.

% === Retrieval results ==========
\medskip \noindent
\textbf{Effect of retrieval models on RAG.}
% \textbf{Comparison of retrieval models.}
To assess the effect of retrieval models, we first calculate the overall Recall score for all retriever models, considering the first paragraph of each Wikipedia page as the answer-containing document. The results are shown in Table~\ref{tb:ret_res}, demonstrating that DPR and BM25+DPR models significantly outperform BM25 and Contriever across all datasets, while DPR and BM25+DPR perform on par with each other.
Table~\ref{tb:answer_generator_retrievers} presents the QA accuracy of RAG with different retrieval models. Following~\citet{When23Mallen}, we use the top-ranked document for the RAG prompt. The results indicate a direct correlation between the performance of the retrieval model and the overall QA accuracy, underscoring the significant impact of the retrieval model on the effectiveness of the downstream task for both vanilla and fine-tuned LMs. Since the performance of RAG with DPR and BM25+DPR is comparable, we use DPR as the retrieval model for subsequent experiments.

\medskip \noindent
\textbf{Analysis of RAG and FT per popularity.}
Figure \ref{fig:all_pop_flant5} illustrates QA accuracy across different popularity buckets, providing insight into the effectiveness of FT and RAG (using Ideal retriever). It is evident that RAG significantly increases accuracy for the least popular entities, which aligns with the findings of \citet{When23Mallen}. Additionally, these figures demonstrate that FT enhances QA accuracy across all popularity levels, with the most notable improvements observed in the least popular buckets for Mistral and Llama3.% certain LMs, such as FlanT5-small, FlanT5-base, and FlanT5-XXL.

\begin{figure}[t]
    \centering
    \includegraphics[width=0.48\textwidth]{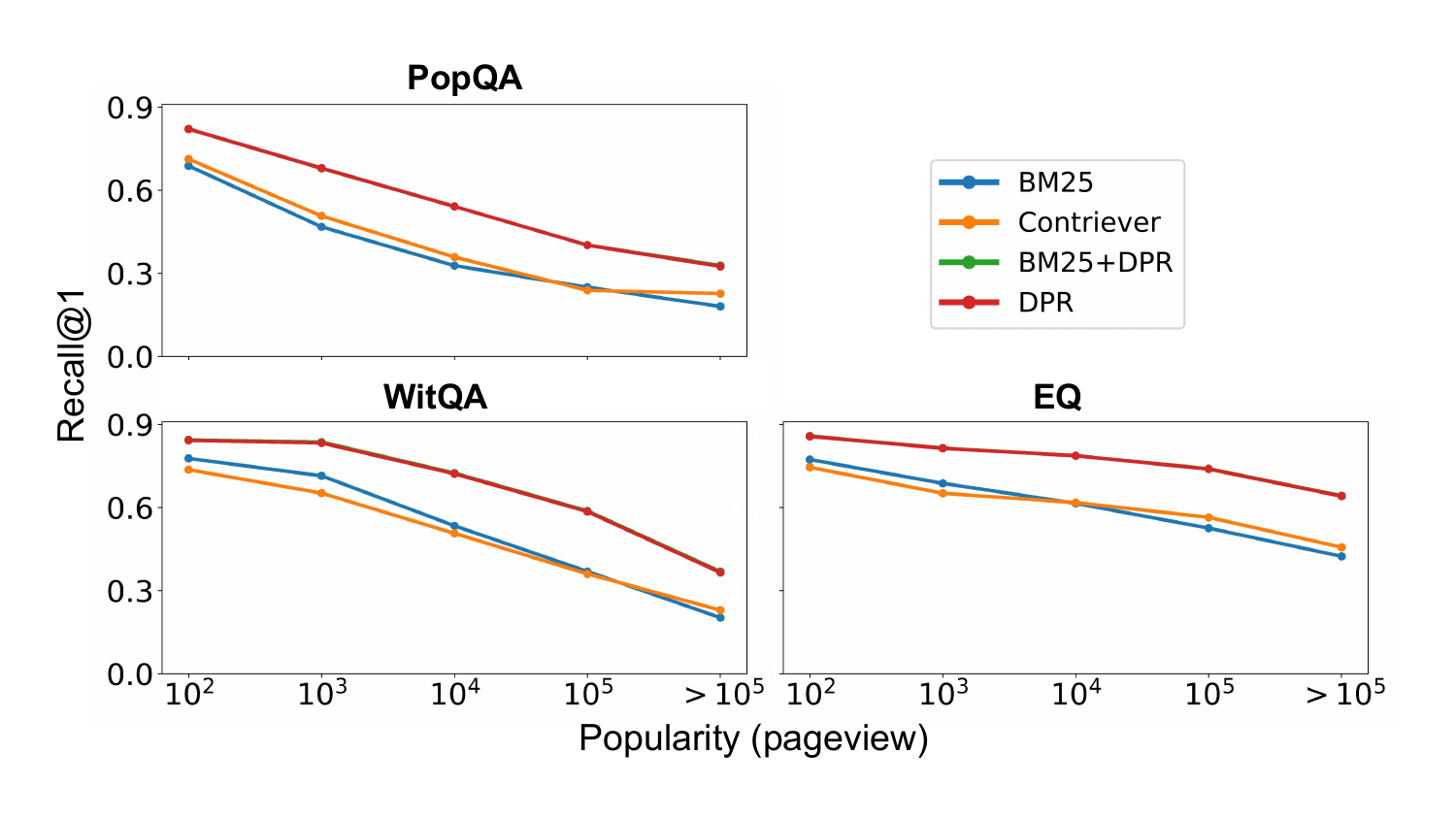}
    \shrink
    \shrink
    \caption{\label{fig:ret_recall} Recall@1 for retrieval models across different popularity levels shows that retrievers perform more effectively with less popular knowledge compared to more popular ones.}
    \shrink
\end{figure}

\medskip \noindent
\textbf{Analysis of retrieval models per popularity.}
Figure~\ref{fig:ret_recall} compares the performance of retrieval models against the Ideal retriever across different popularity buckets. 
The results indicate that retrieval effectiveness is higher for less popular entities compared to more popular entities. This is likely due to the limited occurrences of noisy documents for less popular entities.

\begin{figure}[t]
  \centering
  \includegraphics[width=0.45\textwidth]{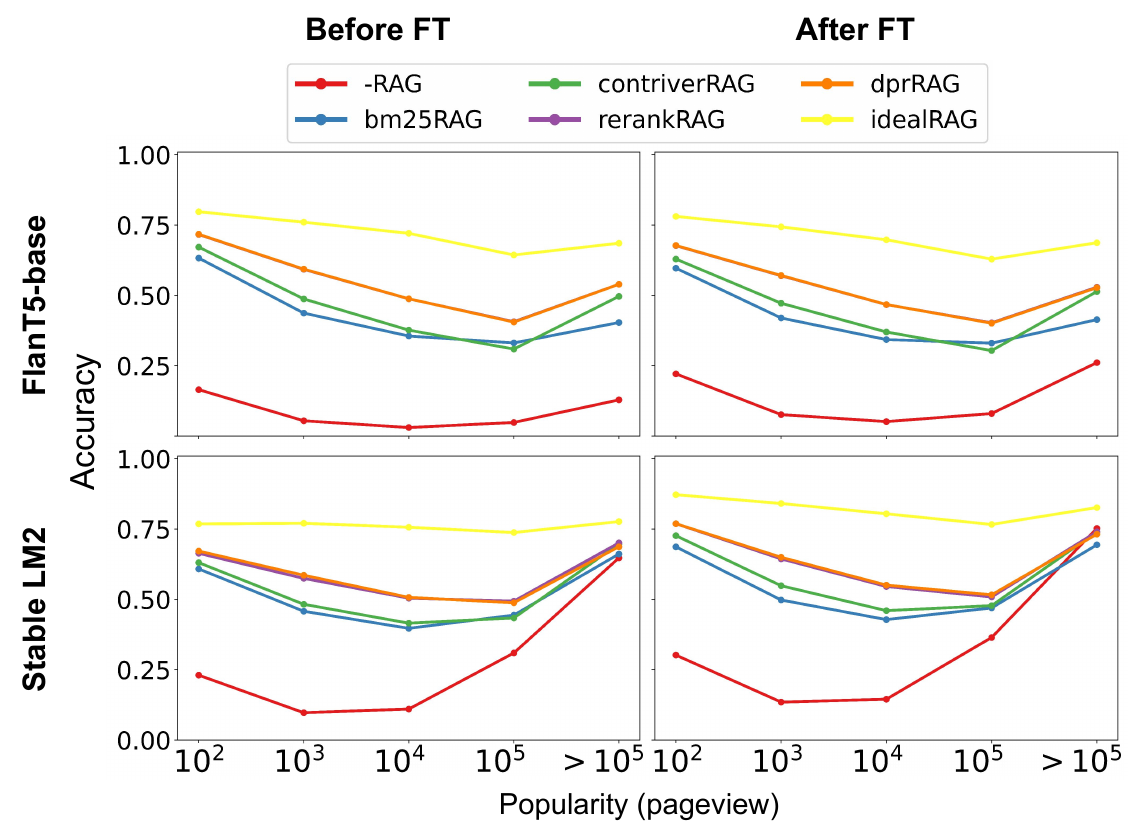}
  \shrink
  \caption{\label{fig:ret_res_bk} Performance of the answer generator task across popularity buckets on \textsc{PopQA}. FT does not alter the overall pattern. Accuracy decreases as the retriever models' performance drops from the least popular bucket to the fourth bucket. Interestingly, accuracy increases for the most popular bucket, indicating that LMs rely on their embedded information for popular entities.
  }
  \shrink
\end{figure}

\noindent
Figure~\ref{fig:ret_res_bk} shows the QA system's accuracy using various retrieval models within the RAG framework across different popularity buckets for the FlanT5-base and StableLM2 models. The left figures display results for vanilla LMs, while the right figures show results for fine-tuned LMs. FT does not alter the pattern of accuracy across popularity buckets but shifts the overall accuracy higher. Interestingly, the accuracy decreases from the less popular bucket to the fourth popularity bucket across different retrievers but increases in the most popular bucket. This reduction in accuracy can be interpreted by the finding in Figure~\ref{fig:ret_recall}, which shows that the retriever's performance decreases as popularity increases. However, it appears that for popular entities, the LMs can ignore noisy information in the input prompt and rely on their internal knowledge to answer questions, resulting in a sudden increase in accuracy despite the retriever's lower performance in the most popular bucket.

% == Different passage number ========
\renewcommand{\arraystretch}{1.25}
\begin{table}[t]
    \small
    \centering
    \caption{The effect of increasing the number of documents in RAG. A significant jump in accuracy is observed when the number of documents is increased to three. However, adding five documents does not significantly affect accuracy. Statistically significant differences in the \textit{(3D)} columns are compared with the \textit{(1D)} and \textit{(5D)} columns. Superscripts (a) and (b) indicate statistically significant differences (better or worse) compared to \textit{(1D)} and \textit{(5D)}, respectively, as determined by the Wilcoxon test (p-value < $0.01$).}
    \shrink
    \label{tb:more_passages_results}
    \begin{tabular}{
    l@{~}|l@{~~}|lll|lll
    % @{~}l@{~~} | c@{~~~} | l@{~~~}l@{~~~}l@{~~~}|l@{~~~}l@{~~~}l@{~}
    }
    \hline
    \textbf{~~} & &
    \multicolumn{3}{c|}{\textbf{-FT+RAG}} &
    \multicolumn{3}{c}{\textbf{+FT+RAG}} \\
    \textbf{Model} & &
    \textbf{(1D)} & \textbf{(3D)} & \textbf{(5D)} &
     \textbf{(1D)} & \textbf{(3D)} & \textbf{(5D)} \\
    \hline

    \multicolumn{8}{c}{\textbf{\textsc{PopQA}}} \\
    \hline

    \multirow{2}{*}{FlanT5-base}
    & DPR & 
    53.36 & 56.67\textsuperscript{(a)} & 56.67 &
    52.20 & 53.46\textsuperscript{(a)} & 53.46 \\
    & Ideal &
    74.50 & 73.06\textsuperscript{(b)} & 75.53 &
    71.34 & 72.02 & 71.75 \\ 
    % \hline

    \multirow{2}{*}{StableLM2}
    & DPR &
    55.74 & 63.98\textsuperscript{(a)} & 64.00 &
    61.44 & 65.33\textsuperscript{(a)} & 65.33 \\
    & Ideal & 
    76.14 & 80.82\textsuperscript{(a)} & 80.68 &
    82.09 & 82.98 & 82.99 \\ 
    % \hline
    
    \multirow{2}{*}{Mistral}
    & DPR &
    60.09 & 65.22\textsuperscript{(a)} & 65.22 &
    61.23 & 63.63\textsuperscript{(a)} & 63.63 \\
    & Ideal &
    80.25 & 81.58\textsuperscript{(a)} & 81.43 &
    78.44 & 80.30\textsuperscript{(a)} & 80.44 \\ 
    % \hline

    \multirow{2}{*}{Llama3}
    & DPR &
    61.46 & 66.66\textsuperscript{(a,b)} & 67.94 &
    62.10 & 66.61\textsuperscript{(a)} & 66.43 \\
    & Ideal &
    81.29 & 82.58\textsuperscript{(a)} & 83.27 &
    81.54 & 82.58 & 82.63 \\ 
    % \hline

    % ========= EQ ============ 
    \hline
    \multicolumn{8}{c}{\textbf{EQ}} \\ 
    \hline
    \multirow{2}{*}{FlanT5-base}
    & DPR & 
    49.54 & 52.39\textsuperscript{(a)} & 52.05 &
    50.39 & 48.90\textsuperscript{(a)} & 48.02 \\
    & Ideal &
    53.92 & 58.02\textsuperscript{(a)} & 58.18 &
    57.63 & 57.24 & 57.52 \\

    \multirow{2}{*}{StableLM2}
    & DPR &
    54.51 & 63.51\textsuperscript{(a)} & 63.75 &
    61.91 & 64.67\textsuperscript{(a)} & 64.52 \\
    & Ideal & 
    60.72 & 69.16\textsuperscript{(a)} & 69.16 &
    68.82 & 70.74\textsuperscript{(a)} & 71.09 \\

    \multirow{2}{*}{Mistral}
    & DPR & 
    54.51 & 66.85\textsuperscript{(a)} & 67.40 &
    61.28 & 63.92\textsuperscript{(a)} & 64.73 \\
    & Ideal &
    68.60 & 71.38\textsuperscript{(a)} & 71.98 &
    64.94 & 68.60\textsuperscript{(a,b)} & 70.31 \\

    \multirow{2}{*}{Llama3}
    & DPR &
    63.27 & 67.76\textsuperscript{(a)} & 68.41 &
    64.00 & 65.56\textsuperscript{(a)} & 65.45 \\
    & Ideal &
    68.67 & 71.50\textsuperscript{(a)} & 72.46 &
    68.64 & 71.89\textsuperscript{(a)} & 71.81 \\

    \hline
    \end{tabular}
    \shrink
    % \shrink
% \vskip -0.1in
\end{table}

\subsection{Stimulus RAG performance}
% === Effect of number of passages 
The second research question \emph{\textbf{(RQ2)}} concerns whether  our proposed Stimulus RAG method can surpass the performance of fine-tuned models.
To evaluate the effect of the stimulus RAG, we first need to investigate how increasing the number of documents in the input prompt affects the accuracy of LMs. Table~\ref{tb:more_passages_results} presents the results of RAG with top-1, top-3 and top-5 documents, shown as (1D), (3D), and (5D), respectively. It shows that using top-3 documents leads to noticeable accuracy improvements in all cases, both before and after FT. For DPR, these results align with those in Table~\ref{tb:ret_res}, where there is a significant increase from Recall@1 to Recall@3. For Ideal retriever, it is important to note that the Ideal retriever is not 100 percent accurate; for some queries, the answer is found in other paragraphs, not just in the summary paragraph.

Another notable observation is that increasing the number of input documents to five results in either negligible accuracy improvement or a decrease in accuracy.
This occurs despite Table~\ref{fig:ret_recall} showing that Recall@5 for DPR is higher than Recall@3. This observation suggests that as the number of documents increases, some LMs struggle to effectively utilize all the information, making it harder to find the correct answer and leading to misunderstandings. The results indicate that adding more textual information to the input prompt should be done judiciously.

% == Highlighting Results ============
\renewcommand{\arraystretch}{1.25}
\begin{table}[t]
    \small
    \centering
    \caption{
    SRAG performance. By adding the extracted hint to the top of the input prompt, SRAG's performance surpasses other settings. 
    % The best results in each row are shown in \textbf{bold}, and the best results in each row are \underline{underlined}.
    Statistically significant differences in the \textit{SRAG(S)} and \textit{SRAG(D)} columns are compared with the \textit{-FT+RAG} and \textit{+FT+RAG} columns. Superscripts (a) and (b) denote statistically significant differences (better or worse) compared to \textit{-FT+RAG} and \textit{+FT+RAG}, respectively, as determined by the Wilcoxon test (p-value < $0.01$).
    }
    \shrink
    \label{tb:highlighting_results}
    \begin{tabular}{
    l|l|ll|ll
    % l|c|l@{~~~}l@{~~~}|l@{~~~}l@{~~~}
    % @{~}l@{~~} | c@{~~~} | c@{~~~}c@{~~~}:c@{~~~}c@{~}
    }
    \hline
    \textbf{~~} & &
    \textbf{-FT+RAG} & \textbf{+FT+RAG} &
    \multicolumn{2}{c}{\textbf{SRAG}} \\
    \textbf{Model} & &
    \textbf{(3D)} & \textbf{(3D)} &
    \textbf{(S)} & \textbf{(D)} \\
    \hline

    \multicolumn{6}{c}{\textbf{\textsc{PopQA}}} \\
    \hline

    \multirow{2}{*}{FlanT5-base}
    & DPR & 
    56.67 & 53.46 &
    \textbf{57.77}\textsuperscript{(a,b)} & 57.67\textsuperscript{(a,b)} \\
    & Ideal &
    73.06 & 72.02 & 
    75.08\textsuperscript{(a,b)} & \textbf{75.29}\textsuperscript{(a,b)} \\

    \multirow{2}{*}{StableLM2}
    & DPR &
    63.98 & 65.33 & 
    65.48\textsuperscript{(a)} & \textbf{66.01}\textsuperscript{(a)} \\
    & Ideal & 
    80.82 & 82.98 &
    82.83\textsuperscript{(a)} & \textbf{83.18}\textsuperscript{(a)} \\
    
    \multirow{2}{*}{Mistral}
    & DPR &
    65.22 & 63.63 & 
    65.84\textsuperscript{(a,b)} & \textbf{66.04}\textsuperscript{(a,b)} \\
    & Ideal &
    81.58 & 80.30 & 
    81.88\textsuperscript{(b)} & \textbf{82.27}\textsuperscript{(a,b)} \\

    \multirow{2}{*}{Llama3}
    & DPR &
    66.66 & 66.61 & 
    \textbf{67.22} & 67.21  \\
    & Ideal &
    \textbf{82.58} & \textbf{82.58} &
    82.42 & 81.60 \\

    % ========= EQ ============ 
    \hline
    \multicolumn{6}{c}{\textbf{EQ}} \\ 
    \hline
    \multirow{2}{*}{FlanT5-base}
    & DPR & 
    52.39 & 48.90 &
    \textbf{55.31}\textsuperscript{(a,b)} & 55.12\textsuperscript{(a,b)} \\
    & Ideal &
    58.02 & 57.24 & 
    \textbf{60.21}\textsuperscript{(a,b)} &  59.88\textsuperscript{(a,b)} \\

    \multirow{2}{*}{StableLM2}
    & DPR &
    63.51 & 64.67 & 65.97\textsuperscript{(a,b)} & \textbf{66.73}\textsuperscript{(a,b)} \\
    & Ideal & 
    69.16 & 70.74 & 71.38\textsuperscript{(a)} & \textbf{71.73}\textsuperscript{(a,b)} \\

    \multirow{2}{*}{Mistral}
    & DPR & 
    66.85 & 63.92 & 68.54\textsuperscript{(a,b)} & \textbf{68.76}\textsuperscript{(a,b)} \\
    & Ideal &
    71.38 & 68.60 & 72.23\textsuperscript{(a,b)} & \textbf{72.45}\textsuperscript{(a,b)} \\

    \multirow{2}{*}{Llama3}
    & DPR &
    67.76 & 65.56 & 68.31\textsuperscript{(b)} & \textbf{68.38}\textsuperscript{(b)} \\
    & Ideal &
    71.70 & 71.89 & 71.71 & \textbf{71.98} \\

    \hline
    \end{tabular}
    \shrink
    \miniskip
% \vskip -0.2in
\end{table}

Table~\ref{tb:highlighting_results} presents the results of the Stimulus RAG with three documents. It is important to note that the hint sentence and document are derived from the top-3 documents, so no extra information is added to the input prompt. This ensures a fair comparison with RAG using three documents, allowing us to solely evaluate the effectiveness of the highlighting method. 
In all cases, Stimulus RAG without FT achieves higher accuracy than fine-tuned LMs with RAG using top-3 documents. This indicates that guiding LMs with a hint not only improves RAG accuracy before FT but can also surpass the effects of FT. These findings demonstrate that better accuracy can be achieved by designing a more advanced RAG system without the complexities and resource demands of FT.

\section{Discussion and Conclusions}~\label{sec:conclusions}
In this paper, we aimed to determine the most suitable approach for customizing language models (LMs) for less-resourced domains. We examined the effectiveness of retrieval augmented generation (RAG) and fine-tuning (FT) methods, focusing on four key aspects: (i) fine-tuning methods, specifically full fine-tuning versus parameter-efficient fine-tuning (PEFT), (ii) data augmentation techniques,  (iii) the type and size of LMs, including decoder-only versus encoder-decoder models ranging from 80 million to 11 billion parameters, and (iv) the performance of retrieval models.
Our findings reveal several key points. First, PEFT enhances downstream task performance and preserves the reasoning abilities of LMs while incorporating new knowledge. Second, prompt-based QA generation exhibits superior performance in factual QA tasks. Third, a small fine-tuned LM with RAG can perform on par with or even surpass a larger LM model. Additionally, RAG’s performance improves with higher-performing retrievers. Notably, when comparing knowledge injection methods, RAG significantly outperforms FT.
We addressed the cost of fine-tuning by developing Stimulus RAG (SRAG), a novel RAG approach that prompts an LM to generate correct responses based on hints provided in the prompt. This method eliminates the need for extensive fine-tuning, making it a cost-effective solution for enhancing LM performance in less-resourced domains.

\section*{Acknowledgments}
This publication is part of the project LESSEN with project number NWA.1389.20.183 of the research program NWA ORC 2020/21 which is (partly) financed by the Dutch Research Council (NWO).

\bibliographystyle{ACM-Reference-Format}
\balance
\bibliography{main}


\begin{thebibliography}{60}

%%% the Web address.

\ifx \showCODEN    \undefined \def \showCODEN     #1{\unskip}     \fi
\ifx \showDOI      \undefined \def \showDOI       #1{#1}\fi
\ifx \showISBNx    \undefined \def \showISBNx     #1{\unskip}     \fi
\ifx \showISBNxiii \undefined \def \showISBNxiii  #1{\unskip}     \fi
\ifx \showISSN     \undefined \def \showISSN      #1{\unskip}     \fi
\ifx \showLCCN     \undefined \def \showLCCN      #1{\unskip}     \fi
\ifx \shownote     \undefined \def \shownote      #1{#1}          \fi
\ifx \showarticletitle \undefined \def \showarticletitle #1{#1}   \fi
\ifx \showURL      \undefined \def \showURL       {\relax}        \fi
% invisible to TeX
\providecommand\bibfield[2]{#2}
\providecommand\bibinfo[2]{#2}
\providecommand\natexlab[1]{#1}
\providecommand\showeprint[2][]{arXiv:#2}

\bibitem[Abbasiantaeb and Aliannejadi(2024)]%
        {Generate24Abbasiantaeb}
\bibfield{author}{\bibinfo{person}{Zahra Abbasiantaeb} {and} \bibinfo{person}{Mohammad Aliannejadi}.} \bibinfo{year}{2024}\natexlab{}.
\newblock \showarticletitle{Generate then Retrieve: Conversational Response Retrieval Using LLMs as Answer and Query Generators}.
\newblock \bibinfo{journal}{\emph{CoRR}}  \bibinfo{volume}{abs/2403.19302} (\bibinfo{year}{2024}).
\newblock
\showeprint[arXiv]{2403.19302}


\bibitem[Adams(2015)]%
        {adams2015bloom}
\bibfield{author}{\bibinfo{person}{Nancy~E Adams}.} \bibinfo{year}{2015}\natexlab{}.
\newblock \showarticletitle{Bloom’s taxonomy of cognitive learning objectives}.
\newblock \bibinfo{journal}{\emph{Journal of the Medical Library Association: JMLA}} \bibinfo{volume}{103}, \bibinfo{number}{3} (\bibinfo{year}{2015}), \bibinfo{pages}{152}.
\newblock


\bibitem[Alberti et~al\mbox{.}(2019)]%
        {Synthetic19Alberti}
\bibfield{author}{\bibinfo{person}{Chris Alberti}, \bibinfo{person}{Daniel Andor}, \bibinfo{person}{Emily Pitler}, \bibinfo{person}{Jacob Devlin}, {and} \bibinfo{person}{Michael Collins}.} \bibinfo{year}{2019}\natexlab{}.
\newblock \showarticletitle{Synthetic {QA} Corpora Generation with Roundtrip Consistency}. In \bibinfo{booktitle}{\emph{Proceedings of the 57th Conference of the Association for Computational Linguistics, {ACL} 2019}}. \bibinfo{pages}{6168--6173}.
\newblock


\bibitem[Asai et~al\mbox{.}(2023)]%
        {Retrieval23Asai}
\bibfield{author}{\bibinfo{person}{Akari Asai}, \bibinfo{person}{Sewon Min}, \bibinfo{person}{Zexuan Zhong}, {and} \bibinfo{person}{Danqi Chen}.} \bibinfo{year}{2023}\natexlab{}.
\newblock \showarticletitle{Retrieval-based Language Models and Applications}. In \bibinfo{booktitle}{\emph{Proceedings of the 61st Annual Meeting of the Association for Computational Linguistics: Tutorial Abstracts, {ACL} 2023}}. \bibinfo{pages}{41--46}.
\newblock


\bibitem[Asai et~al\mbox{.}(2024a)]%
        {SelfRAG23Asai}
\bibfield{author}{\bibinfo{person}{Akari Asai}, \bibinfo{person}{Zeqiu Wu}, \bibinfo{person}{Yizhong Wang}, \bibinfo{person}{Avirup Sil}, {and} \bibinfo{person}{Hannaneh Hajishirzi}.} \bibinfo{year}{2024}\natexlab{a}.
\newblock \showarticletitle{Self-RAG: Learning to Retrieve, Generate, and Critique through Self-Reflection}. In \bibinfo{booktitle}{\emph{The Twelfth International Conference on Learning Representations, {ICLR} (2024)}}.
\newblock


\bibitem[Asai et~al\mbox{.}(2024b)]%
        {Reliable24Asai}
\bibfield{author}{\bibinfo{person}{Akari Asai}, \bibinfo{person}{Zexuan Zhong}, \bibinfo{person}{Danqi Chen}, \bibinfo{person}{Pang~Wei Koh}, \bibinfo{person}{Luke Zettlemoyer}, \bibinfo{person}{Hannaneh Hajishirzi}, {and} \bibinfo{person}{Wen{-}tau Yih}.} \bibinfo{year}{2024}\natexlab{b}.
\newblock \showarticletitle{Reliable, Adaptable, and Attributable Language Models with Retrieval}.
\newblock \bibinfo{journal}{\emph{CoRR}} (\bibinfo{year}{2024}).
\newblock


\bibitem[Askari et~al\mbox{.}(2023a)]%
        {Askari23Test}
\bibfield{author}{\bibinfo{person}{Arian Askari}, \bibinfo{person}{Mohammad Aliannejadi}, \bibinfo{person}{Evangelos Kanoulas}, {and} \bibinfo{person}{Suzan Verberne}.} \bibinfo{year}{2023}\natexlab{a}.
\newblock \showarticletitle{A Test Collection of Synthetic Documents for Training Rankers: ChatGPT vs. Human Experts}. In \bibinfo{booktitle}{\emph{Proceedings of the 32nd {ACM} International Conference on Information and Knowledge Management, {CIKM} (2023)}}. \bibinfo{pages}{5311--5315}.
\newblock


\bibitem[Askari et~al\mbox{.}(2023b)]%
        {Askari23Expand}
\bibfield{author}{\bibinfo{person}{Arian Askari}, \bibinfo{person}{Mohammad Aliannejadi}, \bibinfo{person}{Chuan Meng}, \bibinfo{person}{Evangelos Kanoulas}, {and} \bibinfo{person}{Suzan Verberne}.} \bibinfo{year}{2023}\natexlab{b}.
\newblock \showarticletitle{Expand, Highlight, Generate: RL-driven Document Generation for Passage Reranking}. In \bibinfo{booktitle}{\emph{Proceedings of the 2023 Conference on Empirical Methods in Natural Language Processing, {EMNLP} (2023)}}. \bibinfo{pages}{10087--10099}.
\newblock


\bibitem[Bellagente et~al\mbox{.}(2024)]%
        {Stablelm224Bellagente}
\bibfield{author}{\bibinfo{person}{Marco Bellagente}, \bibinfo{person}{Jonathan Tow}, \bibinfo{person}{Dakota Mahan}, \bibinfo{person}{Duy Phung}, \bibinfo{person}{Maksym Zhuravinskyi}, \bibinfo{person}{Reshinth Adithyan}, \bibinfo{person}{James Baicoianu}, \bibinfo{person}{Ben Brooks}, \bibinfo{person}{Nathan Cooper}, \bibinfo{person}{Ashish Datta}, \bibinfo{person}{Meng Lee}, \bibinfo{person}{Emad Mostaque}, \bibinfo{person}{Michael Pieler}, \bibinfo{person}{Nikhil Pinnaparaju}, \bibinfo{person}{Paulo Rocha}, \bibinfo{person}{Harry Saini}, \bibinfo{person}{Hannah Teufel}, \bibinfo{person}{Niccol{\'{o}} Zanichelli}, {and} \bibinfo{person}{Carlos Riquelme}.} \bibinfo{year}{2024}\natexlab{}.
\newblock \showarticletitle{Stable {LM} 2 1.6B Technical Report}.
\newblock \bibinfo{journal}{\emph{CoRR}}  \bibinfo{volume}{abs/2402.17834} (\bibinfo{year}{2024}).
\newblock


\bibitem[Chen et~al\mbox{.}(2021)]%
        {Evaluating20Chen}
\bibfield{author}{\bibinfo{person}{Anthony Chen}, \bibinfo{person}{Pallavi Gudipati}, \bibinfo{person}{Shayne Longpre}, \bibinfo{person}{Xiao Ling}, {and} \bibinfo{person}{Sameer Singh}.} \bibinfo{year}{2021}\natexlab{}.
\newblock \showarticletitle{Evaluating Entity Disambiguation and the Role of Popularity in Retrieval-Based {NLP}}. In \bibinfo{booktitle}{\emph{Proceedings of the 59th Annual Meeting of the Association for Computational Linguistics and the 11th International Joint Conference on Natural Language Processing, {ACL/IJCNLP} (2021)}}. \bibinfo{pages}{4472--4485}.
\newblock


\bibitem[Cheng et~al\mbox{.}(2023)]%
        {Lift23Cheng}
\bibfield{author}{\bibinfo{person}{Xin Cheng}, \bibinfo{person}{Di Luo}, \bibinfo{person}{Xiuying Chen}, \bibinfo{person}{Lemao Liu}, \bibinfo{person}{Dongyan Zhao}, {and} \bibinfo{person}{Rui Yan}.} \bibinfo{year}{2023}\natexlab{}.
\newblock \showarticletitle{Lift Yourself Up: Retrieval-augmented Text Generation with Self-Memory}. In \bibinfo{booktitle}{\emph{Proceedings of the Annual Conference on Neural Information Processing Systems, (NeurIPS) 2023}}.
\newblock


\bibitem[Cho et~al\mbox{.}(2020)]%
        {Cho20Better}
\bibfield{author}{\bibinfo{person}{Sangwoo Cho}, \bibinfo{person}{Kaiqiang Song}, \bibinfo{person}{Chen Li}, \bibinfo{person}{Dong Yu}, \bibinfo{person}{Hassan Foroosh}, {and} \bibinfo{person}{Fei Liu}.} \bibinfo{year}{2020}\natexlab{}.
\newblock \showarticletitle{Better Highlighting: Creating Sub-Sentence Summary Highlights}. In \bibinfo{booktitle}{\emph{Proceedings of the 2020 Conference on Empirical Methods in Natural Language Processing, {EMNLP} (2020)}}. \bibinfo{pages}{6282--6300}.
\newblock


\bibitem[Chowdhery et~al\mbox{.}(2023)]%
        {PaLM23Chowdhery}
\bibfield{author}{\bibinfo{person}{Aakanksha Chowdhery}, \bibinfo{person}{Sharan Narang}, \bibinfo{person}{Jacob Devlin}, \bibinfo{person}{Maarten Bosma}, \bibinfo{person}{Gaurav Mishra}, \bibinfo{person}{Adam Roberts}, \bibinfo{person}{Paul Barham}, \bibinfo{person}{Hyung~Won Chung}, \bibinfo{person}{Charles Sutton}, \bibinfo{person}{Sebastian Gehrmann}, {and} \bibinfo{person}{et al.}} \bibinfo{year}{2023}\natexlab{}.
\newblock \showarticletitle{PaLM: Scaling Language Modeling with Pathways}.
\newblock \bibinfo{journal}{\emph{Journal of Machine Learning Research}}  \bibinfo{volume}{24} (\bibinfo{year}{2023}), \bibinfo{pages}{240:1--240:113}.
\newblock


\bibitem[Chung et~al\mbox{.}(2024)]%
        {Scaling22Chung}
\bibfield{author}{\bibinfo{person}{Hyung~Won Chung}, \bibinfo{person}{Le Hou}, \bibinfo{person}{Shayne Longpre}, \bibinfo{person}{Barret Zoph}, \bibinfo{person}{Yi Tay}, \bibinfo{person}{William Fedus}, \bibinfo{person}{Eric Li}, \bibinfo{person}{Xuezhi Wang}, \bibinfo{person}{Mostafa Dehghani}, \bibinfo{person}{Siddhartha Brahma}, \bibinfo{person}{Albert Webson}, \bibinfo{person}{Shixiang~Shane Gu}, \bibinfo{person}{Zhuyun Dai}, \bibinfo{person}{Mirac Suzgun}, \bibinfo{person}{Xinyun Chen}, \bibinfo{person}{Aakanksha Chowdhery}, \bibinfo{person}{Sharan Narang}, \bibinfo{person}{Gaurav Mishra}, \bibinfo{person}{Adams Yu}, \bibinfo{person}{Vincent~Y. Zhao}, \bibinfo{person}{Yanping Huang}, \bibinfo{person}{Andrew~M. Dai}, \bibinfo{person}{Hongkun Yu}, \bibinfo{person}{Slav Petrov}, \bibinfo{person}{Ed~H. Chi}, \bibinfo{person}{Jeff Dean}, \bibinfo{person}{Jacob Devlin}, \bibinfo{person}{Adam Roberts}, \bibinfo{person}{Denny Zhou}, \bibinfo{person}{Quoc~V. Le}, {and} \bibinfo{person}{Jason Wei}.}
  \bibinfo{year}{2024}\natexlab{}.
\newblock \showarticletitle{Scaling Instruction-Finetuned Language Models}.
\newblock \bibinfo{journal}{\emph{Journal of Machine Learning Research}}  \bibinfo{volume}{25} (\bibinfo{year}{2024}), \bibinfo{pages}{70:1--70:53}.
\newblock


\bibitem[Cuconasu et~al\mbox{.}(2024)]%
        {Noise2024Florin}
\bibfield{author}{\bibinfo{person}{Florin Cuconasu}, \bibinfo{person}{Giovanni Trappolini}, \bibinfo{person}{Federico Siciliano}, \bibinfo{person}{Simone Filice}, \bibinfo{person}{Cesare Campagnano}, \bibinfo{person}{Yoelle Maarek}, \bibinfo{person}{Nicola Tonellotto}, {and} \bibinfo{person}{Fabrizio Silvestri}.} \bibinfo{year}{2024}\natexlab{}.
\newblock \showarticletitle{The Power of Noise: Redefining Retrieval for {RAG} Systems}. In \bibinfo{booktitle}{\emph{Proceedings of the 47th International {ACM} {SIGIR} Conference on Research and Development in Information Retrieval, {SIGIR} (2024)}}. \bibinfo{pages}{719--729}.
\newblock


\bibitem[{De Cao} et~al\mbox{.}(2021)]%
        {decao2021autoregressive}
\bibfield{author}{\bibinfo{person}{Nicola {De Cao}}, \bibinfo{person}{Gautier Izacard}, \bibinfo{person}{Sebastian Riedel}, {and} \bibinfo{person}{Fabio Petroni}.} \bibinfo{year}{2021}\natexlab{}.
\newblock \showarticletitle{Autoregressive Entity Retrieval}. In \bibinfo{booktitle}{\emph{9th International Conference on Learning Representations, {ICLR} (2021)}}.
\newblock


\bibitem[de~Luis~Balaguer et~al\mbox{.}(2024)]%
        {RAG24Balaguer}
\bibfield{author}{\bibinfo{person}{Maria~Angels de Luis~Balaguer}, \bibinfo{person}{Vinamra Benara}, \bibinfo{person}{Renato~Luiz de Freitas~Cunha}, \bibinfo{person}{Roberto de M.~Estev{\~{a}}o~Filho}, \bibinfo{person}{Todd Hendry}, \bibinfo{person}{Daniel Holstein}, \bibinfo{person}{Jennifer Marsman}, \bibinfo{person}{Nick Mecklenburg}, \bibinfo{person}{Sara Malvar}, \bibinfo{person}{Leonardo~O. Nunes}, \bibinfo{person}{Rafael Padilha}, \bibinfo{person}{Morris Sharp}, \bibinfo{person}{Bruno Silva}, \bibinfo{person}{Swati Sharma}, \bibinfo{person}{Vijay Aski}, {and} \bibinfo{person}{Ranveer Chandra}.} \bibinfo{year}{2024}\natexlab{}.
\newblock \showarticletitle{{RAG} vs Fine-tuning: Pipelines, Tradeoffs, and a Case Study on Agriculture}.
\newblock   \bibinfo{volume}{abs/2401.08406} (\bibinfo{year}{2024}).
\newblock


\bibitem[Dettmers et~al\mbox{.}(2023)]%
        {QLoRA23Dettmers}
\bibfield{author}{\bibinfo{person}{Tim Dettmers}, \bibinfo{person}{Artidoro Pagnoni}, \bibinfo{person}{Ari Holtzman}, {and} \bibinfo{person}{Luke Zettlemoyer}.} \bibinfo{year}{2023}\natexlab{}.
\newblock \showarticletitle{QLoRA: Efficient Finetuning of Quantized LLMs}. In \bibinfo{booktitle}{\emph{Advances in Neural Information Processing Systems 36: Annual Conference on Neural Information Processing Systems, NeurIPS (2023)}}.
\newblock


\bibitem[Gerritse et~al\mbox{.}(2022)]%
        {Gerritse:2022:EMBERT}
\bibfield{author}{\bibinfo{person}{Emma~J Gerritse}, \bibinfo{person}{Faegheh Hasibi}, {and} \bibinfo{person}{Arjen~P de Vries}.} \bibinfo{year}{2022}\natexlab{}.
\newblock \showarticletitle{{Entity-Aware Transformers for Entity Search}}. In \bibinfo{booktitle}{\emph{Proceedings of the 45th International ACM SIGIR Conference on Research and Development in Information Retrieval}} \emph{(\bibinfo{series}{SIGIR '22})}. \bibinfo{pages}{1455--1465}.
\newblock


\bibitem[Godbole and Jia(2023)]%
        {Benchmarking23Godbole}
\bibfield{author}{\bibinfo{person}{Ameya Godbole} {and} \bibinfo{person}{Robin Jia}.} \bibinfo{year}{2023}\natexlab{}.
\newblock \showarticletitle{Benchmarking Long-tail Generalization with Likelihood Splits}. In \bibinfo{booktitle}{\emph{Findings of the Association for Computational Linguistics: {EACL}}}. \bibinfo{pages}{933--953}.
\newblock


\bibitem[Hu et~al\mbox{.}(2024)]%
        {MiniCPM24Hu}
\bibfield{author}{\bibinfo{person}{Shengding Hu}, \bibinfo{person}{Yuge Tu}, \bibinfo{person}{Xu Han}, \bibinfo{person}{Chaoqun He}, \bibinfo{person}{Ganqu Cui}, \bibinfo{person}{Xiang Long}, \bibinfo{person}{Zhi Zheng}, \bibinfo{person}{Yewei Fang}, \bibinfo{person}{Yuxiang Huang}, \bibinfo{person}{Weilin Zhao}, \bibinfo{person}{Xinrong Zhang}, \bibinfo{person}{Zhen~Leng Thai}, \bibinfo{person}{Kai Zhang}, \bibinfo{person}{Chongyi Wang}, \bibinfo{person}{Yuan Yao}, \bibinfo{person}{Chenyang Zhao}, \bibinfo{person}{Jie Zhou}, \bibinfo{person}{Jie Cai}, \bibinfo{person}{Zhongwu Zhai}, \bibinfo{person}{Ning Ding}, \bibinfo{person}{Chao Jia}, \bibinfo{person}{Guoyang Zeng}, \bibinfo{person}{Dahai Li}, \bibinfo{person}{Zhiyuan Liu}, {and} \bibinfo{person}{Maosong Sun}.} \bibinfo{year}{2024}\natexlab{}.
\newblock \showarticletitle{MiniCPM: Unveiling the Potential of Small Language Models with Scalable Training Strategies}.
\newblock   \bibinfo{volume}{abs/2404.06395} (\bibinfo{year}{2024}).
\newblock


\bibitem[Izacard et~al\mbox{.}(2022)]%
        {Unsupervised22Izacard}
\bibfield{author}{\bibinfo{person}{Gautier Izacard}, \bibinfo{person}{Mathilde Caron}, \bibinfo{person}{Lucas Hosseini}, \bibinfo{person}{Sebastian Riedel}, \bibinfo{person}{Piotr Bojanowski}, \bibinfo{person}{Armand Joulin}, {and} \bibinfo{person}{Edouard Grave}.} \bibinfo{year}{2022}\natexlab{}.
\newblock \showarticletitle{Unsupervised Dense Information Retrieval with Contrastive Learning}.
\newblock \bibinfo{journal}{\emph{Trans. Mach. Learn. Res.}} (\bibinfo{year}{2022}).
\newblock


\bibitem[Jeong et~al\mbox{.}(2024)]%
        {Jeong24Adaptive}
\bibfield{author}{\bibinfo{person}{Soyeong Jeong}, \bibinfo{person}{Jinheon Baek}, \bibinfo{person}{Sukmin Cho}, \bibinfo{person}{Sung~Ju Hwang}, {and} \bibinfo{person}{Jong Park}.} \bibinfo{year}{2024}\natexlab{}.
\newblock \showarticletitle{Adaptive-RAG: Learning to Adapt Retrieval-Augmented Large Language Models through Question Complexity}. In \bibinfo{booktitle}{\emph{Proceedings of the 2024 Conference of the North American Chapter of the Association for Computational Linguistics: Human Language Technologies, {NAACL} 2024}}. \bibinfo{pages}{7036--7050}.
\newblock


\bibitem[Jiang et~al\mbox{.}(2023)]%
        {Mistral23Jiang}
\bibfield{author}{\bibinfo{person}{Albert~Q. Jiang}, \bibinfo{person}{Alexandre Sablayrolles}, \bibinfo{person}{Arthur Mensch}, \bibinfo{person}{Chris Bamford}, \bibinfo{person}{Devendra~Singh Chaplot}, \bibinfo{person}{Diego de Las~Casas}, \bibinfo{person}{Florian Bressand}, \bibinfo{person}{Gianna Lengyel}, \bibinfo{person}{Guillaume Lample}, \bibinfo{person}{Lucile Saulnier}, \bibinfo{person}{L{\'{e}}lio~Renard Lavaud}, \bibinfo{person}{Marie{-}Anne Lachaux}, \bibinfo{person}{Pierre Stock}, \bibinfo{person}{Teven~Le Scao}, \bibinfo{person}{Thibaut Lavril}, \bibinfo{person}{Thomas Wang}, \bibinfo{person}{Timoth{\'{e}}e Lacroix}, {and} \bibinfo{person}{William~El Sayed}.} \bibinfo{year}{2023}\natexlab{}.
\newblock \showarticletitle{Mistral 7B}.
\newblock \bibinfo{journal}{\emph{CoRR}} (\bibinfo{year}{2023}).
\newblock
\showeprint[arXiv]{2310.06825}


\bibitem[Kaddour et~al\mbox{.}(2023)]%
        {Challenges23Kaddour}
\bibfield{author}{\bibinfo{person}{Jean Kaddour}, \bibinfo{person}{Joshua Harris}, \bibinfo{person}{Maximilian Mozes}, \bibinfo{person}{Herbie Bradley}, \bibinfo{person}{Roberta Raileanu}, {and} \bibinfo{person}{Robert McHardy}.} \bibinfo{year}{2023}\natexlab{}.
\newblock \showarticletitle{Challenges and Applications of Large Language Models}.
\newblock  (\bibinfo{year}{2023}).
\newblock


\bibitem[Kamphuis et~al\mbox{.}(2023)]%
        {Kamphuis:2023:MMEAD}
\bibfield{author}{\bibinfo{person}{Chris Kamphuis}, \bibinfo{person}{Aileen Lin}, \bibinfo{person}{Siwen Yang}, \bibinfo{person}{Jimmy Lin}, \bibinfo{person}{Arjen~P de Vries}, {and} \bibinfo{person}{Faegheh Hasibi}.} \bibinfo{year}{2023}\natexlab{}.
\newblock \showarticletitle{{MMEAD: MS MARCO Entity Annotations and Disambiguations}}. In \bibinfo{booktitle}{\emph{Proceedings of the 46th International ACM SIGIR Conference on Research and Development in Information Retrieval}} \emph{(\bibinfo{series}{SIGIR '23})}. \bibinfo{pages}{2817--2825}.
\newblock


\bibitem[Kandpal et~al\mbox{.}(2023)]%
        {Large23Kandpal}
\bibfield{author}{\bibinfo{person}{Nikhil Kandpal}, \bibinfo{person}{Haikang Deng}, \bibinfo{person}{Adam Roberts}, \bibinfo{person}{Eric Wallace}, {and} \bibinfo{person}{Colin Raffel}.} \bibinfo{year}{2023}\natexlab{}.
\newblock \showarticletitle{Large Language Models Struggle to Learn Long-Tail Knowledge}. In \bibinfo{booktitle}{\emph{International Conference on Machine Learning, {ICML}, (2023)}}. \bibinfo{pages}{15696--15707}.
\newblock


\bibitem[Karpukhin et~al\mbox{.}(2020)]%
        {Dense20Karpukhin}
\bibfield{author}{\bibinfo{person}{Vladimir Karpukhin}, \bibinfo{person}{Barlas Oguz}, \bibinfo{person}{Sewon Min}, \bibinfo{person}{Patrick S.~H. Lewis}, \bibinfo{person}{Ledell Wu}, \bibinfo{person}{Sergey Edunov}, \bibinfo{person}{Danqi Chen}, {and} \bibinfo{person}{Wen{-}tau Yih}.} \bibinfo{year}{2020}\natexlab{}.
\newblock \showarticletitle{Dense Passage Retrieval for Open-Domain Question Answering}. In \bibinfo{booktitle}{\emph{Proceedings of the 2020 Conference on Empirical Methods in Natural Language Processing, {EMNLP} (2020)}}. \bibinfo{pages}{6769--6781}.
\newblock


\bibitem[Kasai et~al\mbox{.}(2023)]%
        {RealTime22Kasai}
\bibfield{author}{\bibinfo{person}{Jungo Kasai}, \bibinfo{person}{Keisuke Sakaguchi}, \bibinfo{person}{Yoichi Takahashi}, \bibinfo{person}{Ronan~Le Bras}, \bibinfo{person}{Akari Asai}, \bibinfo{person}{Xinyan Yu}, \bibinfo{person}{Dragomir Radev}, \bibinfo{person}{Noah~A. Smith}, \bibinfo{person}{Yejin Choi}, {and} \bibinfo{person}{Kentaro Inui}.} \bibinfo{year}{2023}\natexlab{}.
\newblock \showarticletitle{RealTime {QA:} What's the Answer Right Now?}. In \bibinfo{booktitle}{\emph{Advances in Neural Information Processing Systems 36: Annual Conference on Neural Information Processing Systems, NeurIPS (2023)}}.
\newblock


\bibitem[Lewis et~al\mbox{.}(2019)]%
        {Unsupervised19Lewis}
\bibfield{author}{\bibinfo{person}{Patrick S.~H. Lewis}, \bibinfo{person}{Ludovic Denoyer}, {and} \bibinfo{person}{Sebastian Riedel}.} \bibinfo{year}{2019}\natexlab{}.
\newblock \showarticletitle{Unsupervised Question Answering by Cloze Translation}. In \bibinfo{booktitle}{\emph{Proceedings of the 57th Conference of the Association for Computational Linguistics, {ACL} (2019)}}. \bibinfo{pages}{4896--4910}.
\newblock


\bibitem[Lewis et~al\mbox{.}(2020)]%
        {Retrieval20Patrick}
\bibfield{author}{\bibinfo{person}{Patrick S.~H. Lewis}, \bibinfo{person}{Ethan Perez}, \bibinfo{person}{Aleksandra Piktus}, \bibinfo{person}{Fabio Petroni}, \bibinfo{person}{Vladimir Karpukhin}, \bibinfo{person}{Naman Goyal}, \bibinfo{person}{Heinrich K{\"{u}}ttler}, \bibinfo{person}{Mike Lewis}, \bibinfo{person}{Wen{-}tau Yih}, \bibinfo{person}{Tim Rockt{\"{a}}schel}, \bibinfo{person}{Sebastian Riedel}, {and} \bibinfo{person}{Douwe Kiela}.} \bibinfo{year}{2020}\natexlab{}.
\newblock \showarticletitle{Retrieval-Augmented Generation for Knowledge-Intensive {NLP} Tasks}. In \bibinfo{booktitle}{\emph{Advances in Neural Information Processing Systems 33: Annual Conference on Neural Information Processing Systems 2020, NeurIPS (2020)}}.
\newblock


\bibitem[Lewis et~al\mbox{.}(2021)]%
        {PAQ21Lewis}
\bibfield{author}{\bibinfo{person}{Patrick S.~H. Lewis}, \bibinfo{person}{Yuxiang Wu}, \bibinfo{person}{Linqing Liu}, \bibinfo{person}{Pasquale Minervini}, \bibinfo{person}{Heinrich K{\"{u}}ttler}, \bibinfo{person}{Aleksandra Piktus}, \bibinfo{person}{Pontus Stenetorp}, {and} \bibinfo{person}{Sebastian Riedel}.} \bibinfo{year}{2021}\natexlab{}.
\newblock \showarticletitle{{PAQ:} 65 Million Probably-Asked Questions and What You Can Do With Them}.
\newblock \bibinfo{journal}{\emph{Trans. Assoc. Comput. Linguistics}} (\bibinfo{year}{2021}), \bibinfo{pages}{1098--1115}.
\newblock


\bibitem[Li et~al\mbox{.}(2023)]%
        {Li23Guiding}
\bibfield{author}{\bibinfo{person}{Zekun Li}, \bibinfo{person}{Baolin Peng}, \bibinfo{person}{Pengcheng He}, \bibinfo{person}{Michel Galley}, \bibinfo{person}{Jianfeng Gao}, {and} \bibinfo{person}{Xifeng Yan}.} \bibinfo{year}{2023}\natexlab{}.
\newblock \showarticletitle{Guiding Large Language Models via Directional Stimulus Prompting}. In \bibinfo{booktitle}{\emph{Advances in Neural Information Processing Systems 36: Annual Conference on Neural Information Processing Systems 2023, NeurIPS (2023)}}.
\newblock


\bibitem[Lin et~al\mbox{.}(2021)]%
        {Lin2021Pretrained}
\bibfield{author}{\bibinfo{person}{Jimmy Lin}, \bibinfo{person}{Rodrigo~Frassetto Nogueira}, {and} \bibinfo{person}{Andrew Yates}.} \bibinfo{year}{2021}\natexlab{}.
\newblock \bibinfo{booktitle}{\emph{Pretrained Transformers for Text Ranking: {BERT} and Beyond}}.
\newblock \bibinfo{publisher}{Morgan {\&} Claypool Publishers}.
\newblock


\bibitem[Liu et~al\mbox{.}(2022)]%
        {Few22Haokun}
\bibfield{author}{\bibinfo{person}{Haokun Liu}, \bibinfo{person}{Derek Tam}, \bibinfo{person}{Mohammed Muqeeth}, \bibinfo{person}{Jay Mohta}, \bibinfo{person}{Tenghao Huang}, \bibinfo{person}{Mohit Bansal}, {and} \bibinfo{person}{Colin Raffel}.} \bibinfo{year}{2022}\natexlab{}.
\newblock \showarticletitle{Few-Shot Parameter-Efficient Fine-Tuning is Better and Cheaper than In-Context Learning}. In \bibinfo{booktitle}{\emph{Advances in Neural Information Processing Systems 35: Annual Conference on Neural Information Processing Systems 2022, NeurIPS (2022)}}.
\newblock


\bibitem[Liu et~al\mbox{.}(2024)]%
        {Lost2023Nelson}
\bibfield{author}{\bibinfo{person}{Nelson~F. Liu}, \bibinfo{person}{Kevin Lin}, \bibinfo{person}{John Hewitt}, \bibinfo{person}{Ashwin Paranjape}, \bibinfo{person}{Michele Bevilacqua}, \bibinfo{person}{Fabio Petroni}, {and} \bibinfo{person}{Percy Liang}.} \bibinfo{year}{2024}\natexlab{}.
\newblock \showarticletitle{Lost in the Middle: How Language Models Use Long Contexts}.
\newblock \bibinfo{journal}{\emph{Trans. Assoc. Comput. Linguistics}} (\bibinfo{year}{2024}), \bibinfo{pages}{157--173}.
\newblock


\bibitem[Ma et~al\mbox{.}(2024)]%
        {fine23Ma}
\bibfield{author}{\bibinfo{person}{Xueguang Ma}, \bibinfo{person}{Liang Wang}, \bibinfo{person}{Nan Yang}, \bibinfo{person}{Furu Wei}, {and} \bibinfo{person}{Jimmy Lin}.} \bibinfo{year}{2024}\natexlab{}.
\newblock \showarticletitle{Fine-Tuning LLaMA for Multi-Stage Text Retrieval}. In \bibinfo{booktitle}{\emph{Proceedings of the 47th International {ACM} {SIGIR} Conference on Research and Development in Information Retrieval, {SIGIR} 2024}}. \bibinfo{pages}{2421--2425}.
\newblock


\bibitem[Maekawa et~al\mbox{.}(2024)]%
        {maekawa24witqa}
\bibfield{author}{\bibinfo{person}{Seiji Maekawa}, \bibinfo{person}{Hayate Iso}, \bibinfo{person}{Sairam Gurajada}, {and} \bibinfo{person}{Nikita Bhutani}.} \bibinfo{year}{2024}\natexlab{}.
\newblock \showarticletitle{Retrieval Helps or Hurts? {A} Deeper Dive into the Efficacy of Retrieval Augmentation to Language Models}. In \bibinfo{booktitle}{\emph{Proceedings of the 2024 Conference of the North American Chapter of the Association for Computational Linguistics: Human Language Technologies, {NAACL} 2024}}. \bibinfo{pages}{5506--5521}.
\newblock


\bibitem[Mallen et~al\mbox{.}(2023)]%
        {When23Mallen}
\bibfield{author}{\bibinfo{person}{Alex Mallen}, \bibinfo{person}{Akari Asai}, \bibinfo{person}{Victor Zhong}, \bibinfo{person}{Rajarshi Das}, \bibinfo{person}{Daniel Khashabi}, {and} \bibinfo{person}{Hannaneh Hajishirzi}.} \bibinfo{year}{2023}\natexlab{}.
\newblock \showarticletitle{When Not to Trust Language Models: Investigating Effectiveness of Parametric and Non-Parametric Memories}. In \bibinfo{booktitle}{\emph{Proceedings of the 61st Annual Meeting of the Association for Computational Linguistics {ACL}, (2023)}}. \bibinfo{pages}{9802--9822}.
\newblock


\bibitem[Min et~al\mbox{.}(2023)]%
        {Nonparametric23Min}
\bibfield{author}{\bibinfo{person}{Sewon Min}, \bibinfo{person}{Weijia Shi}, \bibinfo{person}{Mike Lewis}, \bibinfo{person}{Xilun Chen}, \bibinfo{person}{Wen{-}tau Yih}, \bibinfo{person}{Hannaneh Hajishirzi}, {and} \bibinfo{person}{Luke Zettlemoyer}.} \bibinfo{year}{2023}\natexlab{}.
\newblock \showarticletitle{Nonparametric Masked Language Modeling}. In \bibinfo{booktitle}{\emph{Findings of the Association for Computational Linguistics: {ACL} 2023}}. \bibinfo{pages}{2097--2118}.
\newblock


\bibitem[Mosbach et~al\mbox{.}(2023)]%
        {Few23Mosbach}
\bibfield{author}{\bibinfo{person}{Marius Mosbach}, \bibinfo{person}{Tiago Pimentel}, \bibinfo{person}{Shauli Ravfogel}, \bibinfo{person}{Dietrich Klakow}, {and} \bibinfo{person}{Yanai Elazar}.} \bibinfo{year}{2023}\natexlab{}.
\newblock \showarticletitle{Few-shot Fine-tuning vs. In-context Learning: {A} Fair Comparison and Evaluation}. In \bibinfo{booktitle}{\emph{Findings of the Association for Computational Linguistics: {ACL} (2023)}}. \bibinfo{pages}{12284--12314}.
\newblock


\bibitem[Naveed et~al\mbox{.}(2023)]%
        {Naveed23Comprehensive}
\bibfield{author}{\bibinfo{person}{Humza Naveed}, \bibinfo{person}{Asad~Ullah Khan}, \bibinfo{person}{Shi Qiu}, \bibinfo{person}{Muhammad Saqib}, \bibinfo{person}{Saeed Anwar}, \bibinfo{person}{Muhammad Usman}, \bibinfo{person}{Nick Barnes}, {and} \bibinfo{person}{Ajmal Mian}.} \bibinfo{year}{2023}\natexlab{}.
\newblock \showarticletitle{A Comprehensive Overview of Large Language Models}.
\newblock   \bibinfo{volume}{abs/2307.06435} (\bibinfo{year}{2023}).
\newblock


\bibitem[Ovadia et~al\mbox{.}(2023)]%
        {Fine23Ovadia}
\bibfield{author}{\bibinfo{person}{Oded Ovadia}, \bibinfo{person}{Menachem Brief}, \bibinfo{person}{Moshik Mishaeli}, {and} \bibinfo{person}{Oren Elisha}.} \bibinfo{year}{2023}\natexlab{}.
\newblock \showarticletitle{Fine-Tuning or Retrieval? Comparing Knowledge Injection in LLMs}.
\newblock \bibinfo{journal}{\emph{CoRR}}  \bibinfo{volume}{abs/2312.05934} (\bibinfo{year}{2023}).
\newblock
\showeprint[arXiv]{2312.05934}


\bibitem[Robertson and Walker(1994)]%
        {Robertson94bm25}
\bibfield{author}{\bibinfo{person}{Stephen~E. Robertson} {and} \bibinfo{person}{Steve Walker}.} \bibinfo{year}{1994}\natexlab{}.
\newblock \showarticletitle{Some Simple Effective Approximations to the 2-Poisson Model for Probabilistic Weighted Retrieval}. In \bibinfo{booktitle}{\emph{Proceedings of the 17th Annual International {ACM-SIGIR} Conference on Research and Development in Information Retrieval}}. \bibinfo{pages}{232--241}.
\newblock


\bibitem[Robertson and Zaragoza(2009)]%
        {BM2509Robertson}
\bibfield{author}{\bibinfo{person}{Stephen~E. Robertson} {and} \bibinfo{person}{Hugo Zaragoza}.} \bibinfo{year}{2009}\natexlab{}.
\newblock \showarticletitle{The Probabilistic Relevance Framework: {BM25} and Beyond}.
\newblock \bibinfo{journal}{\emph{Found. Trends Inf. Retr.}} \bibinfo{volume}{3}, \bibinfo{number}{4} (\bibinfo{year}{2009}), \bibinfo{pages}{333--389}.
\newblock


\bibitem[Sciavolino et~al\mbox{.}(2021)]%
        {Simple21Sciavolino}
\bibfield{author}{\bibinfo{person}{Christopher Sciavolino}, \bibinfo{person}{Zexuan Zhong}, \bibinfo{person}{Jinhyuk Lee}, {and} \bibinfo{person}{Danqi Chen}.} \bibinfo{year}{2021}\natexlab{}.
\newblock \showarticletitle{Simple Entity-Centric Questions Challenge Dense Retrievers}. In \bibinfo{booktitle}{\emph{Proceedings of the 2021 Conference on Empirical Methods in Natural Language Processing, {EMNLP} 2021}}. \bibinfo{pages}{6138--6148}.
\newblock


\bibitem[Shuster et~al\mbox{.}(2021)]%
        {Retrieval21Shuster}
\bibfield{author}{\bibinfo{person}{Kurt Shuster}, \bibinfo{person}{Spencer Poff}, \bibinfo{person}{Moya Chen}, \bibinfo{person}{Douwe Kiela}, {and} \bibinfo{person}{Jason Weston}.} \bibinfo{year}{2021}\natexlab{}.
\newblock \showarticletitle{Retrieval Augmentation Reduces Hallucination in Conversation}. In \bibinfo{booktitle}{\emph{Findings of the Association for Computational Linguistics: {EMNLP} (2021)}}. \bibinfo{pages}{3784--3803}.
\newblock


\bibitem[Soudani et~al\mbox{.}(2023)]%
        {Data23soudani}
\bibfield{author}{\bibinfo{person}{Heydar Soudani}, \bibinfo{person}{Evangelos Kanoulas}, {and} \bibinfo{person}{Faegheh Hasibi}.} \bibinfo{year}{2023}\natexlab{}.
\newblock \showarticletitle{Data Augmentation for Conversational {AI}}. In \bibinfo{booktitle}{\emph{Proceedings of the 32nd {ACM} International Conference on Information and Knowledge Management, {CIKM} 2023}}. \bibinfo{pages}{5220--5223}.
\newblock


\bibitem[Soudani et~al\mbox{.}(2024a)]%
        {Augmentation24soudani}
\bibfield{author}{\bibinfo{person}{Heydar Soudani}, \bibinfo{person}{Roxana Petcu}, \bibinfo{person}{Evangelos Kanoulas}, {and} \bibinfo{person}{Faegheh Hasibi}.} \bibinfo{year}{2024}\natexlab{a}.
\newblock \showarticletitle{Data Augmentation for Conversational {AI}}. In \bibinfo{booktitle}{\emph{Companion Proceedings of the {ACM} on Web Conference 2024, {WWW} 2024}}, \bibfield{editor}{\bibinfo{person}{Tat{-}Seng Chua}, \bibinfo{person}{Chong{-}Wah Ngo}, \bibinfo{person}{Roy~Ka{-}Wei Lee}, \bibinfo{person}{Ravi Kumar}, {and} \bibinfo{person}{Hady~W. Lauw}} (Eds.). \bibinfo{publisher}{{ACM}}, \bibinfo{pages}{1234--1237}.
\newblock


\bibitem[Soudani et~al\mbox{.}(2024b)]%
        {survey24soudani}
\bibfield{author}{\bibinfo{person}{Heydar Soudani}, \bibinfo{person}{Roxana Petcu}, \bibinfo{person}{Evangelos Kanoulas}, {and} \bibinfo{person}{Faegheh Hasibi}.} \bibinfo{year}{2024}\natexlab{b}.
\newblock \showarticletitle{A Survey on Recent Advances in Conversational Data Generation}.
\newblock \bibinfo{journal}{\emph{arXiv preprint arXiv:2405.13003}} (\bibinfo{year}{2024}).
\newblock


\bibitem[Sun et~al\mbox{.}(2024)]%
        {Head23Kai}
\bibfield{author}{\bibinfo{person}{Kai Sun}, \bibinfo{person}{Yifan~Ethan Xu}, \bibinfo{person}{Hanwen Zha}, \bibinfo{person}{Yue Liu}, {and} \bibinfo{person}{Xin~Luna Dong}.} \bibinfo{year}{2024}\natexlab{}.
\newblock \showarticletitle{Head-to-Tail: How Knowledgeable are Large Language Models (LLMs)? {A.K.A.} Will LLMs Replace Knowledge Graphs?}. In \bibinfo{booktitle}{\emph{Proceedings of the 2024 Conference of the North American Chapter of the Association for Computational Linguistics: Human Language Technologies, {NAACL} (2024)}}. \bibinfo{pages}{311--325}.
\newblock


\bibitem[Thakur et~al\mbox{.}(2021)]%
        {BEIR21Thakur}
\bibfield{author}{\bibinfo{person}{Nandan Thakur}, \bibinfo{person}{Nils Reimers}, \bibinfo{person}{Andreas R{\"{u}}ckl{\'{e}}}, \bibinfo{person}{Abhishek Srivastava}, {and} \bibinfo{person}{Iryna Gurevych}.} \bibinfo{year}{2021}\natexlab{}.
\newblock \showarticletitle{{BEIR:} {A} Heterogeneous Benchmark for Zero-shot Evaluation of Information Retrieval Models}. In \bibinfo{booktitle}{\emph{Proceedings of the Neural Information Processing Systems Track on Datasets and Benchmarks 1, NeurIPS Datasets and Benchmarks 2021}}.
\newblock


\bibitem[Touvron et~al\mbox{.}(2023)]%
        {llama223Touvron}
\bibfield{author}{\bibinfo{person}{Hugo Touvron}, \bibinfo{person}{Louis Martin}, \bibinfo{person}{Kevin Stone}, \bibinfo{person}{Peter Albert}, \bibinfo{person}{Amjad Almahairi}, \bibinfo{person}{Yasmine Babaei}, \bibinfo{person}{Nikolay Bashlykov}, \bibinfo{person}{Soumya Batra}, \bibinfo{person}{Prajjwal Bhargava}, \bibinfo{person}{Shruti Bhosale}, \bibinfo{person}{Dan Bikel}, \bibinfo{person}{Lukas Blecher}, \bibinfo{person}{Cristian Canton{-}Ferrer}, \bibinfo{person}{Moya Chen}, \bibinfo{person}{Guillem Cucurull}, \bibinfo{person}{David Esiobu}, \bibinfo{person}{Jude Fernandes}, \bibinfo{person}{Jeremy Fu}, \bibinfo{person}{Wenyin Fu}, \bibinfo{person}{Brian Fuller}, \bibinfo{person}{Cynthia Gao}, \bibinfo{person}{Vedanuj Goswami}, \bibinfo{person}{Naman Goyal}, \bibinfo{person}{Anthony Hartshorn}, \bibinfo{person}{Saghar Hosseini}, \bibinfo{person}{Rui Hou}, \bibinfo{person}{Hakan Inan}, \bibinfo{person}{Marcin Kardas}, \bibinfo{person}{Viktor Kerkez}, \bibinfo{person}{Madian Khabsa},
  \bibinfo{person}{Isabel Kloumann}, \bibinfo{person}{Artem Korenev}, \bibinfo{person}{Punit~Singh Koura}, \bibinfo{person}{Marie{-}Anne Lachaux}, \bibinfo{person}{Thibaut Lavril}, \bibinfo{person}{Jenya Lee}, \bibinfo{person}{Diana Liskovich}, \bibinfo{person}{Yinghai Lu}, \bibinfo{person}{Yuning Mao}, \bibinfo{person}{Xavier Martinet}, \bibinfo{person}{Todor Mihaylov}, \bibinfo{person}{Pushkar Mishra}, \bibinfo{person}{Igor Molybog}, \bibinfo{person}{Yixin Nie}, \bibinfo{person}{Andrew Poulton}, \bibinfo{person}{Jeremy Reizenstein}, \bibinfo{person}{Rashi Rungta}, \bibinfo{person}{Kalyan Saladi}, \bibinfo{person}{Alan Schelten}, \bibinfo{person}{Ruan Silva}, \bibinfo{person}{Eric~Michael Smith}, \bibinfo{person}{Ranjan Subramanian}, \bibinfo{person}{Xiaoqing~Ellen Tan}, \bibinfo{person}{Binh Tang}, \bibinfo{person}{Ross Taylor}, \bibinfo{person}{Adina Williams}, \bibinfo{person}{Jian~Xiang Kuan}, \bibinfo{person}{Puxin Xu}, \bibinfo{person}{Zheng Yan}, \bibinfo{person}{Iliyan Zarov}, \bibinfo{person}{Yuchen
  Zhang}, \bibinfo{person}{Angela Fan}, \bibinfo{person}{Melanie Kambadur}, \bibinfo{person}{Sharan Narang}, \bibinfo{person}{Aur{\'{e}}lien Rodriguez}, \bibinfo{person}{Robert Stojnic}, \bibinfo{person}{Sergey Edunov}, {and} \bibinfo{person}{Thomas Scialom}.} \bibinfo{year}{2023}\natexlab{}.
\newblock \showarticletitle{Llama 2: Open Foundation and Fine-Tuned Chat Models}.
\newblock \bibinfo{journal}{\emph{CoRR}}  \bibinfo{volume}{abs/2307.09288} (\bibinfo{year}{2023}).
\newblock
\showeprint[arXiv]{2307.09288}


\bibitem[Tunstall et~al\mbox{.}(2023)]%
        {Zephyr23Tunstall}
\bibfield{author}{\bibinfo{person}{Lewis Tunstall}, \bibinfo{person}{Edward Beeching}, \bibinfo{person}{Nathan Lambert}, \bibinfo{person}{Nazneen Rajani}, \bibinfo{person}{Kashif Rasul}, \bibinfo{person}{Younes Belkada}, \bibinfo{person}{Shengyi Huang}, \bibinfo{person}{Leandro von Werra}, \bibinfo{person}{Cl{\'{e}}mentine Fourrier}, \bibinfo{person}{Nathan Habib}, \bibinfo{person}{Nathan Sarrazin}, \bibinfo{person}{Omar Sanseviero}, \bibinfo{person}{Alexander~M. Rush}, {and} \bibinfo{person}{Thomas Wolf}.} \bibinfo{year}{2023}\natexlab{}.
\newblock \showarticletitle{Zephyr: Direct Distillation of {LM} Alignment}.
\newblock \bibinfo{journal}{\emph{CoRR}}  \bibinfo{volume}{abs/2310.16944} (\bibinfo{year}{2023}).
\newblock
\showeprint[arXiv]{2310.16944}


\bibitem[Ushio et~al\mbox{.}(2023)]%
        {Empirical2023ushio}
\bibfield{author}{\bibinfo{person}{Asahi Ushio}, \bibinfo{person}{Fernando Alva-Manchego}, {and} \bibinfo{person}{Jose Camacho-Collados}.} \bibinfo{year}{2023}\natexlab{}.
\newblock \showarticletitle{An Empirical Comparison of {LM}-based Question and Answer Generation Methods}. In \bibinfo{booktitle}{\emph{Findings of the Association for Computational Linguistics: ACL 2023}}. \bibinfo{pages}{14262--14272}.
\newblock


\bibitem[van Hulst et~al\mbox{.}(2020)]%
        {Hulst:2020:REL}
\bibfield{author}{\bibinfo{person}{Johannes~M van Hulst}, \bibinfo{person}{Faegheh Hasibi}, \bibinfo{person}{Koen Dercksen}, \bibinfo{person}{Krisztian Balog}, {and} \bibinfo{person}{Arjen~P de Vries}.} \bibinfo{year}{2020}\natexlab{}.
\newblock \showarticletitle{{REL: An Entity Linker Standing on the Shoulders of Giants}}. In \bibinfo{booktitle}{\emph{Proceedings of the 43rd International ACM SIGIR Conference on Research and Development in Information Retrieval}} \emph{(\bibinfo{series}{SIGIR '20})}. \bibinfo{pages}{2197--2200}.
\newblock


\bibitem[Wei et~al\mbox{.}(2022)]%
        {cot22Wei}
\bibfield{author}{\bibinfo{person}{Jason Wei}, \bibinfo{person}{Xuezhi Wang}, \bibinfo{person}{Dale Schuurmans}, \bibinfo{person}{Maarten Bosma}, \bibinfo{person}{Brian Ichter}, \bibinfo{person}{Fei Xia}, \bibinfo{person}{Ed~H. Chi}, \bibinfo{person}{Quoc~V. Le}, {and} \bibinfo{person}{Denny Zhou}.} \bibinfo{year}{2022}\natexlab{}.
\newblock \showarticletitle{Chain-of-Thought Prompting Elicits Reasoning in Large Language Models}. In \bibinfo{booktitle}{\emph{Advances in Neural Information Processing Systems 35: Annual Conference on Neural Information Processing Systems 2022, NeurIPS (2022)}}.
\newblock


\bibitem[Zaken et~al\mbox{.}(2022)]%
        {BitFit22Zaken}
\bibfield{author}{\bibinfo{person}{Elad~Ben Zaken}, \bibinfo{person}{Yoav Goldberg}, {and} \bibinfo{person}{Shauli Ravfogel}.} \bibinfo{year}{2022}\natexlab{}.
\newblock \showarticletitle{BitFit: Simple Parameter-efficient Fine-tuning for Transformer-based Masked Language-models}. In \bibinfo{booktitle}{\emph{Proceedings of the 60th Annual Meeting of the Association for Computational Linguistics (Volume 2: Short Papers), {ACL} (2022)}}. \bibinfo{pages}{1--9}.
\newblock


\bibitem[Zhang et~al\mbox{.}(2024b)]%
        {tinyllama2024Peiyuan}
\bibfield{author}{\bibinfo{person}{Peiyuan Zhang}, \bibinfo{person}{Guangtao Zeng}, \bibinfo{person}{Tianduo Wang}, {and} \bibinfo{person}{Wei Lu}.} \bibinfo{year}{2024}\natexlab{b}.
\newblock \showarticletitle{TinyLlama: An Open-Source Small Language Model}.
\newblock  (\bibinfo{year}{2024}).
\newblock


\bibitem[Zhang et~al\mbox{.}(2024a)]%
        {Zhang24raft}
\bibfield{author}{\bibinfo{person}{Tianjun Zhang}, \bibinfo{person}{Shishir~G. Patil}, \bibinfo{person}{Naman Jain}, \bibinfo{person}{Sheng Shen}, \bibinfo{person}{Matei Zaharia}, \bibinfo{person}{Ion Stoica}, {and} \bibinfo{person}{Joseph~E. Gonzalez}.} \bibinfo{year}{2024}\natexlab{a}.
\newblock \showarticletitle{{RAFT:} Adapting Language Model to Domain Specific {RAG}}.
\newblock  (\bibinfo{year}{2024}).
\newblock


\end{thebibliography}

% \appendix

\end{document}